\def\year{2022}\relax
\documentclass[letterpaper]{article} % DO NOT CHANGE THIS
\usepackage{aaai22}  % DO NOT CHANGE THIS
\usepackage{times}  % DO NOT CHANGE THIS
\usepackage{helvet}  % DO NOT CHANGE THIS
\usepackage{courier}  % DO NOT CHANGE THIS
\usepackage[hyphens]{url}  % DO NOT CHANGE THIS
\usepackage{graphicx} % DO NOT CHANGE THIS
\usepackage{natbib}  % DO NOT CHANGE THIS AND DO NOT ADD ANY OPTIONS TO IT
\usepackage{caption} % DO NOT CHANGE THIS AND DO NOT ADD ANY OPTIONS TO IT
\DeclareCaptionStyle{ruled}{labelfont=normalfont,labelsep=colon,strut=off} % DO NOT CHANGE THIS
\usepackage[ruled,linesnumbered]{algorithm2e}

\usepackage{amsmath}
\usepackage{amssymb}
\usepackage{amsthm}
\usepackage{subfigure}

\usepackage{booktabs}
\usepackage{multirow}

\usepackage{color}
\usepackage[usenames,dvipsnames]{xcolor}
\usepackage{xcolor}
\usepackage[normalem]{ulem}

\newcommand{\zhou}[1]{{\color{black}#1}}
\newcommand{\zjb}[1]{{\color{black}#1}}

\newcommand{\li}[1]{{\color{black}#1}}
\newcommand{\hide}[1]{} 

\newtheorem{definition}{Definition}

\usepackage{newfloat}
\usepackage{listings}
\title{Adversarial Neural Trip Recommendation}
\author{
    Linlang Jiang\textsuperscript{\rm 1, 2}\thanks{This work was done when the first author was an intern in Baidu Research under the supervision of the second author.},
    Jingbo Zhou\textsuperscript{\rm 2},
    Tong Xu\textsuperscript{\rm 1},
    Yanyan Li\textsuperscript{\rm 2},
    Hao Chen\textsuperscript{\rm 3},
    Jizhou Huang\textsuperscript{\rm 3},
    Hui Xiong\textsuperscript{\rm 4}
}
\affiliations{
\textsuperscript{\rm 1}University of Science and Technology of China, 
\textsuperscript{\rm 2}Business Intelligence Lab, Baidu Research, 
\textsuperscript{\rm 3}Baidu Inc.\\
\textsuperscript{\rm 4}Artificial Intelligence Thrust, The Hong Kong University of Science and Technology\\
linlang@mail.ustc.edu.cn,  \{zhoujingbo, liyanyanliyanyan, chenhao13, huangjizhou01\}@baidu.com\\tongxu@ustc.edu.cn, xionghui@ust.hk
}

\usepackage{bibentry}
\begin{document}

\maketitle

\begin{abstract}
Trip recommender system, which targets at recommending a trip consisting of several ordered  Points of Interest (POIs), has long been treated as an important application for many location-based services. Currently, most prior arts generate trips following pre-defined objectives based on constraint programming, which may fail to reflect the complex latent patterns hidden in the human mobility data. And most of these methods are usually difficult to respond in real time when the number of POIs is large. To that end, we propose an \underline{A}dversarial \underline{N}eural \underline{T}rip Recommendation (ANT) framework to tackle the above challenges. First of all, we devise a novel attention-based encoder-decoder trip generator that can learn the correlations among POIs and generate well-designed trips under given constraints. Another novelty of ANT relies on an adversarial learning strategy integrating with reinforcement learning to guide the trip generator to produce high-quality trips. For this purpose, we introduce a discriminator, which distinguishes the generated trips from real-life trips taken by users, to provide reward signals to optimize the generator. Moreover, we devise a novel pre-train schema based on learning from demonstration, which speeds up the convergence to achieve a sufficient-and-efficient training process. Extensive experiments on four real-world datasets validate the effectiveness and efficiency of our proposed ANT framework, which demonstrates that ANT could remarkably outperform the state-of-the-art baselines with short response time.
\end{abstract}

\section{INTRODUCTION}
Trip recommendation (or trip planning) aims to recommend a trip consisting of several ordered Points of Interest (POIs) for a user to maximize the user experience.
This problem has been extensively investigated \zhou{over the past years} \cite{lim2015personalized, chen2016learning, jia2019joint}.
Most existing \zjb{studies} tackle the problem by a two-stage process.
First, they majorly exploit POI popularity, user preferences, or POI co-occurrence to score POIs and design various objective functions respectively.
Then, they model the trip recommendation problem as a combinatorial problem: Orienteering problem \cite{golden1987orienteering}, and generate trips by maximizing the pre-defined objective with the help of constraint programming (CP).

Though using this \zjb{CP-based} paradigm to solve such a combinatorial problem is very popular over the past years, its drawbacks are still obvious.
First, the recommended trips by such methods are optimized by the pre-defined objective function, which may not follow the latent patterns \zjb{hidden in the human mobility data generated by users.}
For instance, according to the statistics from a real-life trip dataset from Beijing (see Experiment section), after watching a film, 26\% users choose to go to a restaurant while only less than 1\% users choose to go to a Karaoke bar.
However, the pre-defined objective may not be capable of capturing such mobility sequential preferences and generate unusual trips like (cinema $\rightarrow$ Karaoke bar).
Second,
%\li{the time cost of such methods is so high that these methods can not handle hundreds of POIs in a city to response user request in real time, which is shown in our experiment section.}
the time complexity of such CP-based methods is usually too high to handle hundreds of POIs in a city in real time. As shown in our experiment section, the response time of such methods with 100 POIs can be more than 1 minute.
Such a weakness is very disruptive to the user experience.

% \zhou{the time cost of such methods is still very high. As equivalent with Orienteering problem, searching for the optimal solution under pre-defined objective for trip recommendation is NP-hard. Even though these methods can be accelerated by approximation algorithm \li{with} sacrificing effectiveness, they usually cannot handle more than hundreds of POIs in a city like \cite{chen2013automatic,jia2019joint}}. 
% \zhoucom{are there more references for this?}
% Our experiments also show that 
% \zhou{such CP-based methods} cannot \zjb{response in real time if there are} a large number of \zhou{POIs in a city/region.}

To this end, we propose an \underline{A}dversarial \underline{N}eural  \underline{T}rip Recommendation (ANT) framework \zjb{to solve the challenges mentioned above}.
At first, we propose an encoder-decoder based trip generator that can generate the trip under given constraints in an end-to-end fashion.
Concretely, the encoder takes advantage of multi-head self-attention to capture correlations among POIs.
Afterwards, the decoder subsequently selects POI into a trip with mask mechanism to meet the given constraints while maintaining a novel context embedding to represent the contextual environment when choosing POIs.
Second, we 
devise an adversarial learning strategy into the 
specially designed reinforcement learning paradigm to train the generator.
Specifically, we introduce a discriminator to distinguish the real-life trips taken by users from the trips generated by the trip generator for better learning the latent human mobility patterns. During the training process,
once trips are produced by the generator, they will be evaluated by the discriminator while the feedback from the discriminator can be regarded as reward signals to optimize the generator.
Therefore, the generator will push itself to generate high-quality trips to obtain high rewards from the discriminator.
% which in turn guides the trip generator to produce realistic and well-designed trips.
Finally, a significant distinction of our framework from existing trip planning methods is that we do not adopt the traditional constraint programming methodology.
Considering the excellent performance for inference(prediction) of the deep-learning (DL) based models, the efficiency of our method is much better than such CP-based methods.

To sum up, the contributions of this paper can be summarized as follows:
\begin{itemize}
    \item To the best of our knowledge, we are the first to propose an end-to-end DL-based framework to study the trip recommendation problem.
    \item We devise a novel encoder-decoder model to generate trips under given constraints.
    Furthermore, we propose an adversarial learning strategy integrating with reinforcement learning to guide the trip generator to produce trips that follow the latent human mobility patterns.
    \item We conduct extensive experiments on four large-scale real-world datasets.
    The results demonstrate that ANT remarkably outperforms the state-of-the-art techniques from both effectiveness and efficiency perspectives.
\end{itemize}

\section{RELATED WORK}
\zhou{Our study is related with POI recommendation and trip recommendation problems which are briefly discussed in this section respectively.}

\subsection{POI Recommendation}
POI recommendation \zhou{usually} takes the user's historical check-ins as input and aims to \zhou{predict} the POIs that the user is interested in. \zhou{This problem has been extensively investigated in the past years. For example,}
\citet{yang2017bridging} \zhou{proposed to} jointly learn user embeddings and POI embeddings simultaneously to fully comprehend user-POI interactions and predict user preference on POIs under various contexts.
\citet{ma2018point} \zhou{investigated to} utilize attention mechanism to seek what factors of POIs users are concerned about, integrating with geographical influence.
\zhou{\citet{luo2020spatial} studied to build a multi-level POI recommendation model with considering the POIs in different spatial granularity levels.}
However, such methods target on recommending an individual POI not a sequence of POI, and \zhou{do not} consider the dependence and correlations among POIs.
In addition, these methods do not take time budget into consideration while it is vital to recommend trips under the time budget constraint.

\subsection{Trip Recommendation}
Trip recommendation aims to recommend a sequence of POIs (i.e. trip) to maximize user experience under given constraints. 
\citet{lim2015personalized} focused on user interest based on visit duration and personalize the POI visit duration for different users.
\citet{chen2016learning} modeled the POI transit probabilities, integrating with some manually designed features to suggest trips.
\zhou{Another study} modeled POIs and users in a unified latent space by integrating the co-occurrences of POIs, user preferences and POI popularity\cite{jia2019joint}.
These methods above share similar constraints: a start POI, an end POI and a time budget or trip length constraint, and they all maximize respective pre-defined objectives by adopting constraint programming.
However, such pre-defined objectives may fail to generate trips
\zhou{that follow the latent human mobility patterns among POIs.} 
% \zhoucom{revise later} and the efficiency is restricted by the constraint programming solver.
\li{Different from these methods, \citet{gu2020enhancing} focused on the attractiveness of the routes between POIs to recommend trips and generate trips by using greedy algorithm.
However, only modeling users and POIs in category space may not be capable of learning the complex human mobility patterns.}
\zjb{The prediction performance based on greedy strategy is also not satisfied enough.}
% \zhoucom{is there difference of this method from other CP-based methods?}
\section{PRELIMINARIES}
\zhou{In this section, we first introduce the basic concepts and notations, and then we give a formal definition of the trip recommendation problem.}

\subsection{Settings and Concepts}
A \textbf{POI} $l$ is a unique location with geographical coordinates $(\alpha,\beta)$ and a category $c$, i.e. $l=<(\alpha, \beta), c>$. 
%\textbf{Checkin-in}:
A \textbf{check-in} is a record that indicates a user $u$  arrives in a POI $l$ at timestamp $t_a$ and leaves at
timestamp $t_d$, which can be represented as $r = (u, l, t_a, t_d)$. 
We denote all check-ins as $\mathcal{R}$ and the check-ins on \zhou{a} specific location $l$ as $\mathcal{R}_l$. 
% \begin{definition}[Check-in]
% A check-in is a record that indicates a user $u$  arrives in a POI $l$ at timestamp $t_a$ and leaves at
% timestamp $t_d$, which can be represented as $r = (u, l, t_a, t_d)$. 
% We denote all check-ins as $\mathcal{R}$ and the check-ins on \zhou{a} specific location $l$ as $\mathcal{R}_l$.
% \end{definition}
Since we have the check-ins generated by users, we can estimate the user duration time on POIs. Given a POI $l$ and corresponding check-in data $\mathcal{R}_l$, the expected duration time of a user spends on the POI is denoted by $T_d(l)$, which is the average duration time of all check-ins on location $l$:
\begin{equation}
    T_d(l) = \frac {\sum \limits_{(u,l,t_a,t_d) \in \mathcal{R}_l } t_d - t_a} {|\mathcal{R}_l|}
\end{equation}

% \jiang{The transit time a user takes to move from a POI $l_i$ to another POI $l_j$ denoted by $T_e(l_i, l_j)$ can be estimated by the distance between them and the walking speed of the user (e.g. 2m/s). Thus, the time cost along the trip can be calculated by summing all the stay time of each POI and all the time cost on the transit between two successive POIs.} 
\zhou{We denote the transit time from a POI $l_i$ to another POI $l_j$ as $T_e(l_i, l_j)$. 
The time cost along one trip can be calculated by summing all the duration time of each POI and all the time cost on the transit between POIs.
In our experiment, the transit time is estimated by the \li{distance} between POIs and the walking speed of the user (e.g. 2m/s). } 
% \zhoucom{should we mention the type of distance, and other possible methods for estimate the time?}

A \textbf{trip} is an ordered sequence of POIs ${S} = l_0 \rightarrow l_1 \rightarrow \cdots \rightarrow l_n$. 
Given a query user u, a time budget $T_{max}$ and a start POI $l_0$, we aim to plan a trip ${S} = l_0 \rightarrow l_1 \rightarrow \cdots
\rightarrow l_n$ for the user. 
We name the query user, the start POI and the time budget a \textbf{trip query}, denoted as a triple $q = (u, l_0, T_{max})$.
% \begin{definition}[Trip] \label{defi:trip}
% A trip is an ordered sequence of POIs ${S} = l_0 \rightarrow l_1 \rightarrow \cdots \rightarrow l_n$.
% \end{definition}

% \begin{definition}[Query]\label{query}
% Given a query user u, a time budget $T_{max}$ and a start POI $l_0$, we aim to plan a trip ${S} = l_0 \rightarrow l_1 \rightarrow \cdots
% \rightarrow l_n$ for the user. 
% We name the query user, the start POI and the time budget a \textit{trip query}, denoted as a triple $q = (u, l_0, T_{max})$.
% \end{definition}

%\subsection{Trip Planning}\label{prob_def}
\subsection{Trip Recommendation}\label{prob_def}
Now we define the trip recommendation problem formally.
Given a trip query $q=(u, l_0, T_{max})$, we aim to recommend a well-designed trip that does not exceed the time budget and maximize the \zhou{likelihood} that the user will follow the planned trip.
\li{For convenience, we denote the sum of transit time from current POI to the next POI and duration time on the next POI as
$T_a(S_{i}, S_{i+1}) = T_d(S_{i + 1}) + T_e(S_{i}, S_{i+1})$}, $S_i$ is the $i$-th POI in trip $S$.
So the time cost on the planned trip denoted as $T({S})$ can be calculated by 
% $T({S}) = \sum\limits_{i = 1}^{|{S}|} T_d(S_i)
%  + \sum\limits_{i = 0}^{|{S}|-1}  T_e(S_{i}, S_{i+1})$, $S_i$
$T({S}) = T_d({S_0}) +  \sum\limits_{i = 0}^{|{S}|-1}  T_a(S_{i}, S_{i+1})$.
% $T({S}) = \sum\limits_{i = 1}^{|{S}|} T_d(l_{{S}_i})
%  + \sum\limits_{i = 0}^{|{S}|-1}  T_e(l_{{S}_i}, l_{{S}_j})$.
%
% Now we define the trip p, lanning problem formally.
% Given a user $u$, a starting POI $l_0$, the user's time budget $T_{max}$ $\mathcal{A}_u^{l_0}$ consisting of $N$ POIs, $\mathcal{A}_u^{l_0} = \{l_i\}_{0 \le i \le N - 1}$, that is to say, an instance $S$,
% we aim to generate a plan that does not exceeds his/her time budget and maximizes the probability that the user will follow the plan.
% The plan is denoted as ${S} = l_0 \rightarrow l_1
% \rightarrow l_2 \rightarrow \cdots \rightarrow l_n $, where $l_i \in \mathcal{A}_u^{l_0},\ 0 \le i \le n $ and $n$ indicates of number
% of POIs in the plan excluding the starting POI $l_0$.
% Therefore, the time cost on the plan denoted as $T({S})$ can be calculated by $T({S}) = \sum\limits_{i = 1}^s T(l_i)
% + \sum\limits_{i = 0}^{s-1} \sum\limits_{j = 1}^{s} T(d_{ij})$.
% Here the time cost on the plan does not include the time the user spend on POI $l_0$ because we can assume the user is already at POI $l_0$. 
Overall, the problem can be formulated as follows:
\begin{equation}\label{obj}
		\max_{T({S}) \le T_{max}} P({S} \mid q)
\end{equation}

\section{APPROACH}

\begin{figure}[t]
\centering
\includegraphics[width=0.8\columnwidth]{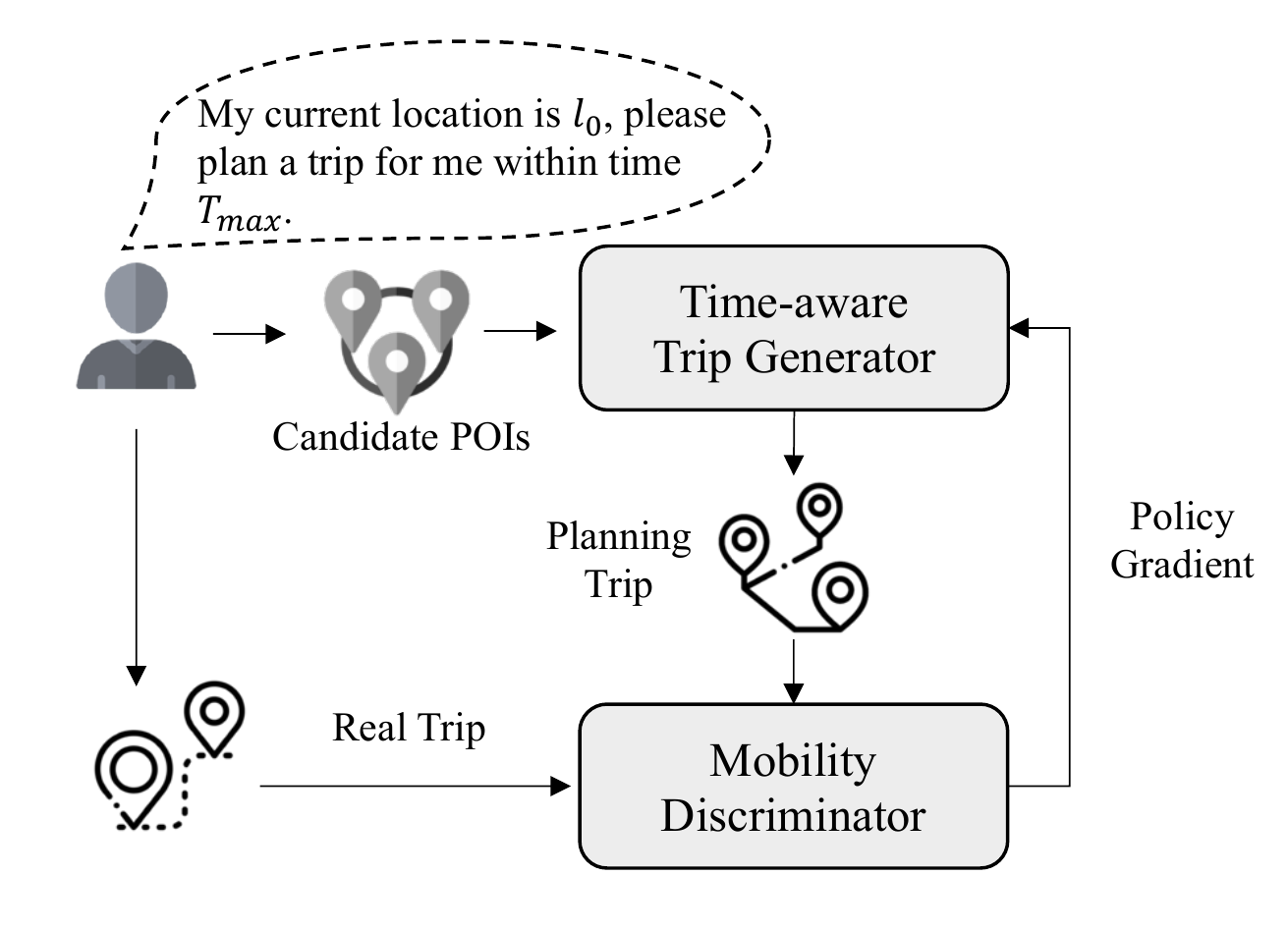}
\vspace{-2mm}
\caption{An overview of the proposed framework.}
\label{fig-overview}
\vspace{-2mm}
\end{figure}

\zhou{The overall framework of ANT is shown in Figure \ref{fig-overview}. We first selectively retrieve hundreds of POIs to construct a candidate set. Next, we use a well-devised novel time-aware trip generator $G$ to generate the well-planned trip for users with incorporating the time budget and the POI correlation.}
\label{reinforce}
%\li{The trip generation process is a sequential decision process, that is to say, at each time step we are supposed to select a POI until we obtain a complete trip.So we model the trip generation procedure as a Markov Decision Process(MDP), where we regard selecting POI as \textbf{action}, the probability distribution on POIs as a stochastic \textbf{policy}, and contextual environment(e.g. available time, selected POIs) when selecting POIs as \textbf{state}.Therefore, our goal is to learn a optimal policy, which guarantees that we can always take the best action, i.e. the POI with highest probability is most promising option.}\li{To obtain a well-designed policy, we follow the Generative Adversarial Network(GAN) structure \cite{goodfellow2014generative} which has a discriminator $D$ to provide the feedback compared with the real trips. Therefore, the generator can be trained through policy gradient to draw the generated trips and the real trips closer.}

\zhou{The trip generation process can be considered as a sequential decision process, that is to say, at each step there is a smart \textbf{agent} to select the best POI which can finally form an optimal trip.
Thus, we model the trip generation procedure as a Markov Decision Process(MDP)\cite{bellman1957markovian}, where we regard selecting POI as \textbf{action}, the probability distribution on POIs \zjb{to be selected} as a stochastic \textbf{policy}, and contextual environment(e.g. available time, selected POIs) when selecting POIs as \textbf{state}.
Therefore, our goal is to learn an optimal policy, which guarantees that the \textbf{agent} can always take the best action, i.e. the POI with the highest probability is the most promising option.
To train the policy, we construct a discriminator $D$ (following the Generative Adversarial Networks(GAN) structure \cite{goodfellow2014generative}) to provide feedback compared with the \zjb{real-life trips taken by users}. Therefore, the generator can be trained through policy gradient \zjb{by reinforcement learning} to draw the generated trips and the real-life trips closer.}

%\zhou{The time-aware trip generator follows the Generative Adversarial Network(GAN) structure \cite{goodfellow2014generative}  which has a discriminator $D$ to provide the feedback compared with the real trips. Therefore, the generator can be trained through policy gradient to draw the generated trips and the real trips closer.}

\subsection{Candidate Construction}
As for a trip query, it is \zhou{usually not necessary} to take all the POIs into consideration to plan a reasonable trip.
For instance, those POIs that are too far away from the start POI are impossible to be part of the trip. 
\zhou{Here} we propose a rule-based retrieval procedure to pick up a small amount of POI \zhou{candidates} from the large POI corpus, named candidate set, which incorporates the impact of connection among trips and geographic influence.

\begin{figure}[t]
\centering
\includegraphics[width=0.6\columnwidth]{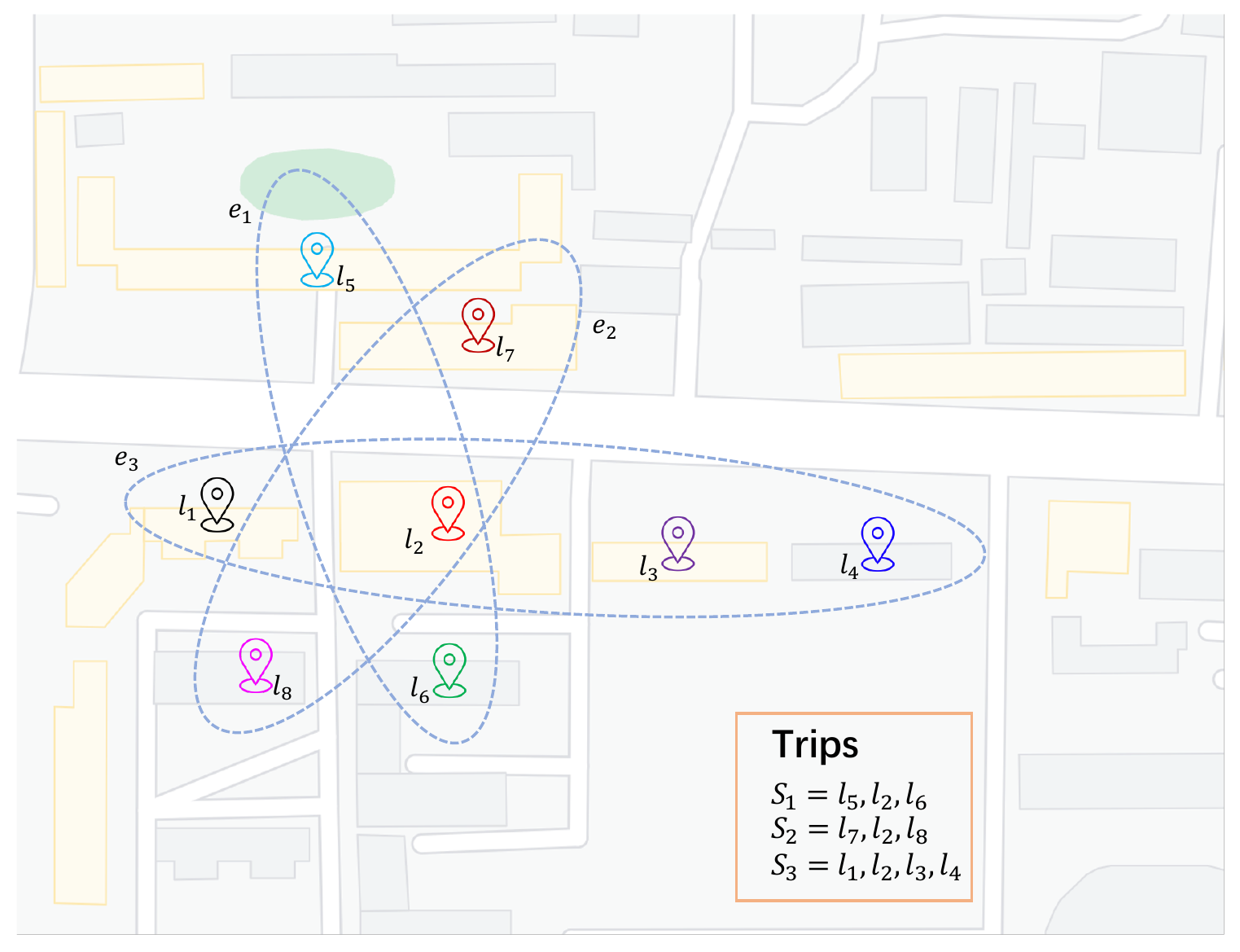}
\caption{An instance of hypergraph construction.}
\label{fig-hypergraph}
\vspace{-4mm}
\end{figure}

\subsubsection{Drawing Lessons from Other Trips}
If a user requests a trip at the start location $l_0$, former trips that are associated with $l_0$ are promising to provide a reference.
Inspired by this, we could assume that given the start POI $l_0$ of a trip query, those POIs that once co-occurred with $l_0$ in the same trip could be potential options for the trip query, which can be named \textit{drawing lessons from other trips}.
Hypergraph provides a natural way to gather POIs belonging to different trips and also to glimpse other trips via hyperedges.
\begin{definition}[Trip Hypergraph]
Let $G=(L,E)$ denote a hypergraph, where $L$ is the vertex set and $E$ is the hyperedge set.
Each vertex represents a POI $l_i$ and each hyperedge $e \in E$ connects two or more vertices, representing a trip.
\end{definition}
Specifically, we use trips in the training set to build the trip hypergraph.
On one hand, all the POIs in the same trip are linked by a hyperedge, which preserves the matching information between POIs and trips.
On the other hand, a POI may exist in arbitrary hyperedges, connecting different trips via hyperedges.
Given a trip query $(u, l_0, T_{max})$, POIs that are connected with $l_0$ via hyperedges are promising to be visited for the upcoming trip request, so we directly add them into the candidate set.
Figure \ref{fig-hypergraph} is a simple example of trip hypergraph retrieval.
If the start POI of the upcoming trip query is $l_2$, POIs $\{l_1, l_3, l_4, l_5, l_6, l_7, l_8\}$ will be added into the candidate set for the corresponding trip query.

\subsubsection{Spatial Retrieval}
Distance between \zhou{users} and POIs is a crucial factor affecting user's decisions in \zhou{location-relative recommendation.}
It is typical that a user's check-ins are mostly centralized on several areas\zhou{\cite{hao2016user,hao2020unified}}, which is the famous geographical clustering phenomenon and is adopted by earlier work to enhance the performance of location recommendation \cite{ma2018point, lian2014geomf, li2015rank}.
% \zhoucom{can we give some citations about this?}
\zhou{Therefore, except for the candidates generated by hypergraph, we also add POIs into candidate set from near to far}.
{In our framework, we generate fixed-length candidate sets for every trip query, denoted as $\mathcal{A}_q$ for the corresponding trip query $q$. We first use the hypergraph to generate candidates and then use the spatial retrieval. In other words, if the numbers of candidates generated by the hypergraph retrieval for different trip queries are smaller than the pre-defined number $|\mathcal{A}_q|$, we pad the candidate set with the sorted POIs} 
\li{in order of} 
\zhou{distance to a fixed length.}
\subsection{Time-aware Trip Generator}\label{generator}
\begin{figure}[t]
\centering
\includegraphics[width=0.8\columnwidth]{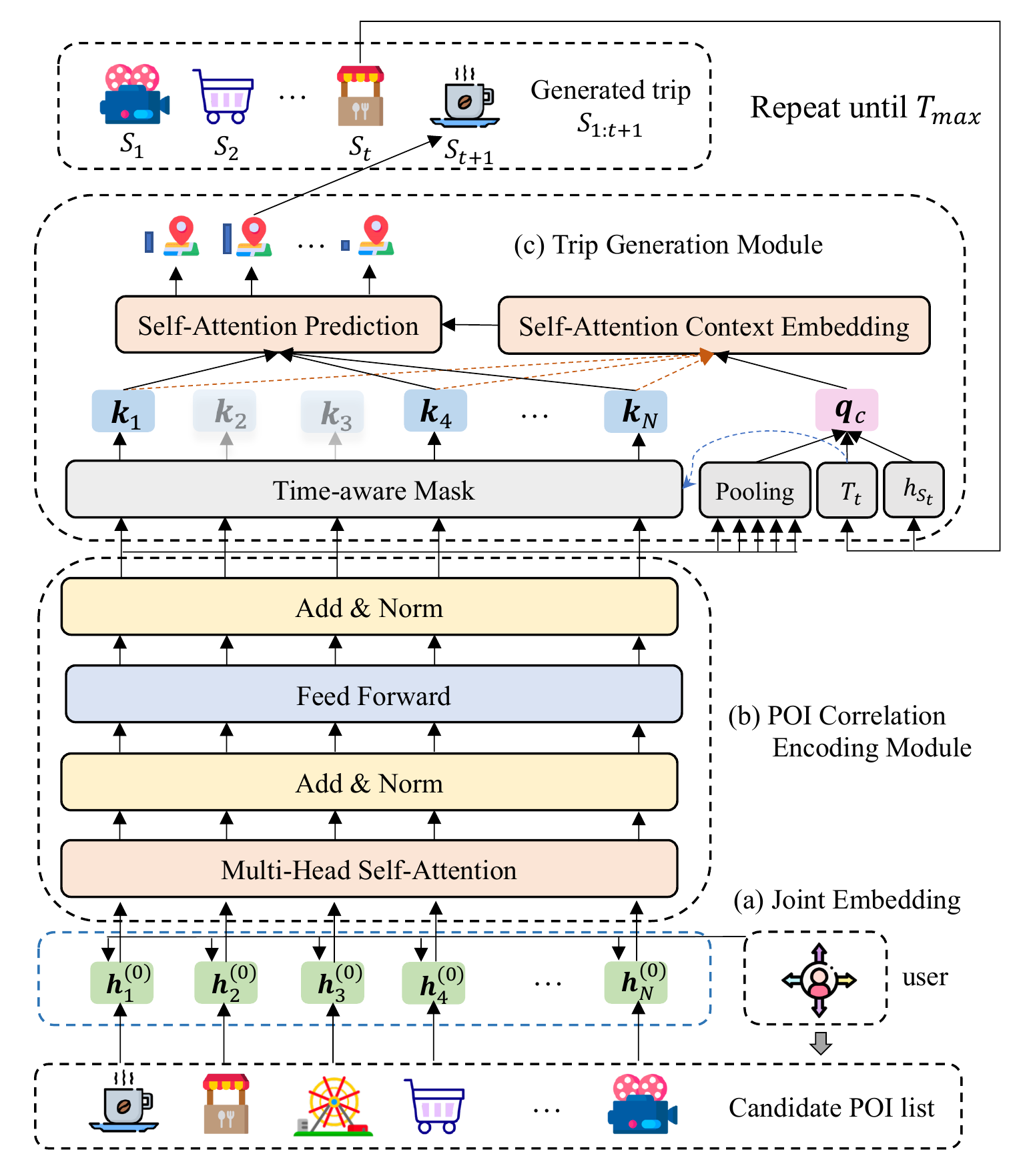}
% \vspace{-1mm}
\caption{Illustration of time-aware trip generator.}
\label{fig-generator}
\vspace{-2mm}
\end{figure}

% \zhou{Here we introduce a novel encoder-decoder framework to incorporate the essential time factor for our trip recommendation problem. }
% The traditional encoder-decoder framework mostly encodes an input sequence to a fixed vector by a recurrent neural network, which is called context vector. Then the decoder decodes the output sequence from another recurrent neural network based on the context vector.
% \zhou{These approaches ignore the essential time factor for our trip recommendation problem. Here we use a time-aware trip generation policy to make personalized trip recommendation for users.
As shown in Figure \ref{fig-generator}, the generator consists of two main components: 1) a POI correlation encoding module (i.e., the encoder), which outputs the representation of all POIs in the candidate set; 2) a trip generation module (i.e., the decoder), which
{selects location sequentially by maintaining a special context embedding, and keeps the time budget constraint satisfied by masking mechanism.}

\subsubsection{Joint Embedding}
Given the trip query $(u, l_0, T_{max})$ and the corresponding selected candidate set $\mathcal{A}_q$, we use a simple linear transform to combine the user $u$ and the POI $l_i$ in $\mathcal{A}_q$ with its category $c$ for embedding:
\begin{equation} 
    h_i^{(0)} =  [x_{l_i}; x_{c}; x_u] \mathbf{W_I} + b_I
\end{equation}
where $x_{l_i}$, $x_{c}$, $x_u$ are POI embedding, category embedding and user embedding (which are all trainable embedding), $[a; b; c]$ means concatenation of vectors $a$, $b$, $c$, and  $\mathbf{W_I}$, $b_I$ are trainable parameters.
Thus, we get the matrix presentation of the candidates $\mathbf{H}^{(0)} \in \mathbb{R}^{N \times d}$, each row of $\mathbf{H}^{(0)}$ is the representation of a POI in the candidate set.

\subsubsection{POI Correlation Encoding}
\zhou{We apply a self-attention encoder to produce the representation of locations. The reasons to use the self-attention encoder can be justified from two perspectives. The first reason is due to the permutation invariance for sets. For a candidate set, \zjb{the order of POIs in this set} is invariant to the final result, i.e. any permutation of the inputs is supposed to produce the same output representation. Thus, we do not adopt the classical RNN-based encoder architecture because it focuses on the sequential information of the inputs, which is not suitable for our problem. Second, a reasonable generated trip is supposed to consider the relationship between POIs.
For instance, after staying at a restaurant for a while a person is more interested in other kinds of POIs but not another restaurant.
So it is helpful to produce a POI representation with attention to other POIs. }

The encoder we apply is similar to the encoder used in the Transformer architecture \cite{vaswani2017attention} while we remove the position encoding, which is not suitable for our problem.
We stack multiple attention layers and each layer has the same sublayers: a multi-head attention(MHA), \zjb{and} a point-wise feed-forward network(FFN).
The initial input of the first attention layer is $\mathbf{H}^{(0)}$
and we apply the scaled dot-product attention for each head in layer $l$ as:
\begin{equation} 
    head_i^{(l)}  = \mathrm{Attn}(\mathbf{H}^{(l - 1)} \mathbf{W}_Q, \mathbf{H}^{(l - 1)} \mathbf{W}_K,
    \mathbf{H}^{(l - 1)} \mathbf{W}_V)
\end{equation}
where $1 \le i \le M, \mathbf{W}_Q, \mathbf{W}_K, \mathbf{W}_V \in \mathbb{R}^ {d \times d_h}, d_h = d / M$, $M$ is the number of heads and $d_h$ is the dimension for each head. The scaled dot-product attention computes as:
\begin{equation} 
    \mathrm{Attn}(\mathbf{Q}, \mathbf{K}, \mathbf{V}) = softmax(\frac{\mathbf{Q} \mathbf{K}^T}{\sqrt{d_h}}) \mathbf{V}
\end{equation}
where the softmax is row-wise. $M$ attention heads are able to capture different aspects of attention information and the results from each head are concatenated followed by a linear projection to get the final output of the MHA. We compute the output of MHA sublayer as:
\begin{equation}
    \hat{\mathbf{H}}^{(l)}  = [head_1^{(l)};  \cdots; head_M^{(l)}] \mathbf{W}_O
\end{equation}
where $\mathbf{W}_O \in \mathbb{R}^{d \times d}$.
\zhou{We endow the encoder with nonlinearity by adding interactions between dimensions by using the FFN sublayer.} The FFN we apply is a two-layer feed-forward network, whose output is computed as:
\begin{equation}
    \mathbf{H}^{(l)} = \mathrm{ReLu} ( \hat{\mathbf{H}}^{(l)} \mathbf{W}^{f1} + \mathbf{b}^{f1}) \mathbf{W}^{f2} + {b}^{f2}
\end{equation}
where $\mathbf{W}^{f1} \in \mathbb{R}^{d \times d_{f}}, \mathbf{W}^{f} \in \mathbb{R}^{d_{f} \times d}$. Note that all 
the parameters for each attention layer is unique.

Besides, to stabilize and speed up converging, the multi-head attention and feed-forward network are both followed by skip connection and batch normalization \cite{vaswani2017attention}.
To sum up, by considering the interactions and inner relationship among POIs, the encoder transforms the original embeddings of POIs into informative representations.
\subsubsection{Trip Generation}
\li{It is of great importance to consider the contextual environment when planning the trip so we design a novel context embedding integrating candidate information, time budget and selected POIs.}

{\bfseries Self-Attention Context Embedding.}
By aggregating the location embeddings, we apply a mean \li{pooling} of final location embedding $\bar{h}^{(L)} = \frac{1}{N} \sum \limits_{i=1} ^ {N} h_i^{(L)}$ as candidate embedding.
During the process of decoding, the decoder selects a POI from the candidate set once at a time based on selected POIs $S_{t^\prime}$, $t^\prime < t$ and the available time left .
We keep track of the remaining available time $T_t$ at time step $t$.
Initially $T_1 = T_{max} - T_d(S_0)$, and $T_t$ is updated as:
\begin{equation} 
    % T_{t + 1} = T_t - T_e(S_{t-1}, S_{t}) - T_d(S_{t}), t \ge 1
    T_{t + 1} = T_t - T_a(S_{t-1}, S_{t}), t \ge 1
\end{equation}
where $S_0 = l_0$.
\zhou{Following existing methods to represent the contextual environment in the procedure of decoding \cite{bello2016neural, kool2018attention}, we employ a novel context embedding $h_c$ \zhou{conditioned on candidate set and remaining time}, which will change along the decoding proceeds.}
The context embedding $h_c$ is defined as:
\begin{equation} 
    h_c = [\bar{h}^{(L)}; h_{S_{t-1}}^{(L)}; T_t], t \ge 1
\end{equation}
where $h_c \in \mathbb{R}^{1 \times (2d+1)}$.

% 在决定下一次选什么POI之前，需要先看一下有哪些POI是可以选的，利用这个信息，我们再一次更新context embedding.
% Before we decide which POI to add into the trip at time step $t$, it is important to look back the information about candidates and \li{figure out} what choices we have.
% So we first glimpse the candidates and compute the refined context embedding with attention to output from the encoder.
% Note that the POIs that are selected before or exceed the time budget are not supposed to be considered here so we mask them.
% We update the context embedding with attention to the output from the encoder.
\li{Before deciding which POI to add into the trip at time step $t$, it is important to look back the information about candidates and remind ourselves which POIs are optional and which POIs should not be considered because they break the given constraints.
Therefore, we first glimpse the candidates that are optional, i.e. are never selected before and do not exceed the time budget, and then integrate the information with attention to the output from the encoder:}
% Concretely, the keys and values come from the location embedding $\mathbf{H}^{(L)}$ from the encoder and the query comes exactly from context embedding:
\begin{equation} 
    q_c  = h_c \mathbf{W}_Q^c  
    \quad k_i = h_i^{(L)} \mathbf{W}_K^c \quad v_i = h_i^{(L)} \mathbf{W}_V^c \\
\end{equation}
% \vspace{-3mm}
\begin{equation} 
    %  \alpha_{tj}  = \frac{\Theta(T_t - T_d(l_j) - T_e(S_{t - 1}, l_j))exp(\frac{q_c^T k_j}{\sqrt{d}})}
    % {\sum\limits_{l_m\in\mathcal{A}_q\backslash S_{0:t-1}} \Theta(T_t - T_d(l_m) - T_e(S_{t-1}, l_m)) exp(\frac{q_c^T k_m}{\sqrt{d}})}
     \alpha_{tj}  = \frac{\Theta(T_t - T_a(S_{t - 1}, l_j))exp(\frac{q_c k_j^T}{\sqrt{d}})}
    {\sum\limits_{l_m\in\mathcal{A}_q\backslash S_{0:t-1}} \Theta(T_t - T_a(S_{t-1}, l_m)) exp(\frac{q_c k_m^T}{\sqrt{d}})}
\end{equation}
where $\mathbf{W}_Q^c \in \mathbb{R}^{(2d+1) \times d}, \mathbf{W}_K^c, \mathbf{W}_V^c \in \mathbb{R}^{d \times d}$, $h_i^{(L)}$ is the $i$-th row of the location embedding matrix $\mathbf{H}^{(L)}$,
and $\Theta(\cdot)$ is a Heaviside step function, which plays a crucial role as the time-aware mask operator. Thus, the refined context embedding $\bar{h}_c$ is computed as:
\begin{equation} 
\bar{h}_c = \sum_{l_j \in \mathcal{A}_q} \alpha_{tj} \cdot v_j
\end{equation}
% Then we compute the compatibility of the query with all the candidates except the POIs that are masked.
% \zhoucom{give more explain and intuition about the equation (11)}
We omit the multi-head due to the page limit.
% The output $\bar{h}_c$ here is similar to glimpse in \cite{bello2016neural}.

{\bfseries Self-Attention Prediction.} After getting the refined context embedding, we apply a final attention layer with a single attention head with mask mechanism.
% We compute the compatibility of the refined context embedding with all locations with mask scheme.
% same as Equation \ref{equation_mask}.
\begin{equation} 
    u_{cj} = 
\begin{cases}
\frac{\bar{h}_c k_j^T}{\sqrt{d}} & \text{otherwise.}\\
% -\infty & \text{if}\  l_j \in S_{0:t - 1} \ \text{or} \ T_t  <  T_d(l_j) + T_e(S_{t -1}, l_j).
-\infty & \text{if}\  l_j \in S_{0:t - 1} \ \text{or} \ T_t  < T_a(S_{t -1}, l_j).
\end{cases}
\end{equation}
\li{Finally, the softmax is applied to get the probability distribution:}
% The compatibility here can be regarded as unnormalized logits and the final probability of choosing a location is computed by using a softmax:
\begin{equation} 
    p(S_t = l_j | \bar{h}_c) = \frac{e^{u_{cj}}} {\sum_{l_m \in \mathcal{A}_q} e^{u_{cm}}}
    \label{equ:prob}
\end{equation}

The decoding proceeds until there is no enough time left and then we get the entire trip generated by the decoder $S_{0:t}$.
To sum up, by maintaining a context embedding and using the representation of location from the encoder, the decoder \li{constructs a trip}
% selects the location sequentially
with attention mechanism and meets the constraints by mask mechanism.
%\zhoucom{revise the following parts later}

%\subsection{Adversarial Learning}
\subsection{Policy Optimization by Adversarial Learning}
\zhou{The next problem is how to train the encoder-decoder framework for trip generation.}
% \jiang{To make the recommended trips more realistic and practical,}
We devise a mobility discriminator to distinguish \zjb{real-life} trips \zjb{taken by users} between generated trips, which provides feedback to guide the optimization of the trip generator. \zhou{After the evaluation between generated trips and real-life trips,  the output of the discriminator can be regarded as reward signals to improve the generator.}
% \jiang{Once trips are produced by the generator, they will be evaluated by the discriminator. After that, the discriminator compares the generated trips with real trips and updates itself while the output of discriminator can be regard as reward signals to optimize the generator.}
% \jiang{During}
\zhou{By} the adversarial procedure, the generator pushes itself to generate high-quality trips to obtain high rewards from the discriminator.
% During the adversarial procedure, the generator could produce high-quality trips that are capable to get great evaluation from the discriminator.
% Understanding latent human mobility patterns is crucial for generating practical and reasonable trips.
% However, the generated trips may fail to follow the latent human mobility patterns.
% Generally, the real trips taken by users could be regarded as optimal trips that follow latent human mobility patterns.
% Thus, we devise a mobility discriminator to distinguish the real trips with generated trips and propose an adversarial learning strategy to optimize the trip generator with the feedback provided by the discriminator.

\subsubsection{Mobility Discriminator}
\zhou{The task for the discriminator essentially is binary classification. Here we apply a simple but effective one-layer Gated Recurrent Unit (GRU) \cite{cho2014learning}, followed by a two-layer feed-forward network to accomplish this task.}
% \jiang{Compared to complicated trip generation procedure, the task for the discriminator is a much more easier binary classification task.
% Therefore, we apply a simple but effective one-layer Gated Recurrent Unit(GRU) \cite{cho2014learning}, followed by a two-layer feed-forward network to balance the training process.}
We denote the mobility discriminator as $D_\phi$ and the trip generator as $G_\phi$, where $\theta$ and $\phi$ represent the parameters of the generator and discriminator respectively.
We denote all the real-life trips as $P_{data}$.
As a binary classification task, we train the discriminator $D_\phi$ as follows:
\begin{equation}
    \mathop{\mathrm{max}}\limits_{\phi}  \mathbb{E}_{\hat{S} \sim P_{data}}[\log D_\phi(\hat{S})] + \mathbb{E}_{S\sim G_\theta}[\log (1 - D_\phi(S))]
\end{equation}
% \vspace{-5mm}
\subsubsection{Adversarial Learning with Policy Gradient}

\zhou{We adopt the reinforcement learning technique to train the generator. The standard training algorithm for GAN does not apply to our framework: the discrete output of the trip generator blocks the gradient back-propagation, making it unable to optimize the generator \cite{yu2017seqgan}. As described previously, the trip generation process is a sequential decision problem, leading us to tackle the problem by adopting reinforcement learning techniques. With modeling the trip generation procedure as an MDP, an important setting is to regard the score from the discriminator as reward.
% We also combined the POIs that have been selected $S_{0:t}$ until time step $t$ to refine context embedding as state, and the output probability as a stochastic policy.
%Since we regard the output from discriminator as reward 
Thus, we define the loss as:
}
$\mathcal{L}(S) =  \mathbb{E}_{p_{\theta}(S \mid q)}
    [D_\phi(S)]$, which represents the expected score for the generated trip $S$ given trip query $q$.
Following REINFORCE \cite{williams1992simple} algorithm, we optimize the loss by gradient ascent:
\begin{equation} 
    \nabla \mathcal{L}(\theta \mid q) = 
      \mathbb{E}_{p_{\theta}(S \mid q)} [D_\phi(S) \nabla \log p_\theta(S \mid q)]
      \label{eqa:policy gradient}
\end{equation}

\subsubsection{Learning from Demonstration}
\zhou{In order to accelerate the training process and further improve the performance, we propose a novel pre-train schema based on learning from demonstration \cite{silver2010learning}, which not only fully utilizes the data of real-life trips but also obtains a decent trip generator before adversarial learning.}
% \jiang{However,}
Learning directly from rewards is sample-inefficient and hard to achieve the promising performance \cite{yu2017seqgan}, \zhou{which is also our reason to introduce the pre-train schema.}
% \jiang{In order to accelerate the training process and further improve the performance, we propose a pre-train schema based on learning from demonstration, which not only fully utilize the data of real trips but also obtain a decent trip generator before adversarial learning.}
During pre-training, 
we use real-life trips taken as ground-truth, regard choosing POI at each time step as a multi-classification problem and optimize by \textit{softmax} loss function.
Nevertheless, during inference, the trip generator needs the preceding POI to select the next POI while we have no access to the true preceding POI \li{in training}, which may lead to cumulative poor decisions \cite{samy2015scheduled}.
To bridge such a gap between training and inference, we select POI by sampling with the probability distribution (defined in Equation \ref{equ:prob}) during training. Finally, the loss can be computed as:
% \vspace{-3mm}
\begin{equation} 
    \mathcal{L}_c = - \sum \limits_{\hat{S} \in P_{data}} \sum \limits_{t=1}^{|\hat{S}|} \log p(\hat{S}_t | S_{0:t-1}; \theta)
    \label{eqa: supervisedloss}
\end{equation}
where $S$ is the actual generated trip during training and $\hat{S}$ is the corresponding real-life trip.

\subsubsection{Teacher Forcing}
The training process is usually unstable by optimizing the generator with Equation \ref{eqa:policy gradient} \cite{li2017adversarial}.
The reason behind this is that once the generator deteriorates in some training batches and the discriminator will recognize the unreasonable trips \li{soon}, then the generator will be lost.
The generator knows the generated trips are not good based on the received rewards from the discriminator, but it does not know how to improve the quality of generated trips.
To alleviate this issue and give the generator more access to real-life trips, after we update the generator with adversarial loss, we also feed the generator real-life trips and update it with supervised loss(Equation \ref{eqa: supervisedloss}) again.

To sum up, we first pre-train the trip generator by leveraging demonstration data.
Afterwards, we alternately update the discriminator and the generator with the respective objective.
During updating the generator, we also feed real-life trips to the generator, regulating the generator from deviation from the demonstration data.
\section{EXPERIMENTS}

\subsection{Experimental Setups}
\subsubsection{Dataset}
We use four real-world POI check-in datasets and Table \ref{tbl-dataset} summarizes the statistics of the four datasets.
% The first two datasets are publicly available datasets from Foursquare \cite{yang2014modeling} and the last two datasets are collected from an online map service provider.

\begin{table}[t]
    \centering
  \begin{tabular}{ccccc}
    \toprule
    \multirow{2}{*}{Dataset} & \multicolumn{2}{c}{Foursquare} & \multicolumn{2}{c}{Map} \\
    \cmidrule(r){2-3}  \cmidrule(r){4-5}
    & NYC & Tokyo & Beijing & Chengdu\\
    \midrule
    \# users          & 796     &  2019  & 22399      & 8869 \\
    \# POIs           & 8619   &  14117 & 13008      & 8914 \\
    \# trips   & 16518   &  58893 & 212758     & 95166 \\
    \bottomrule
  \end{tabular}
  \caption{Dataset statistics.}
  \label{tbl-dataset}
  \vspace{-2mm}
\end{table}

\textbf{Foursquare}\cite{yang2014modeling} This real-world check-in dataset includes check-in data in New York City and Tokyo collected from Foursquare.
We sort the check-ins of a user by timestamp and split them into non-overlapping trips.
If the time interval between two successive check-ins is more than five hours, we split them into two trips.

\textbf{Map} This dataset collects real-world check-ins in Beijing and Chengdu from 1 July 2019 to 30 September 2019 \zhou{from an online map service provider in China}. We consider the check-ins in one day as a trip for a user.

% For both Foursquare and Map datasets,
We remove the trips of which length is less than 3 and we remove the POIs visited by fewer than 5 users as they are outliers in the dataset.
We split the datasets in chronological order, where the former 80 \% for training, the medium 10 \% for validation, and the last 10 \% for testing.
% In the specific personalized POI planning problem, we should generate a POI candidate set first.
% For each instance $(S, I)$, we choose the nearest $K - |S|$ POIs and the POIs in the trajectory as the candidate set, which is similar to the negative sampling in the recommendation system.

\subsubsection{Baselines}
We compare the performance of our proposed method with three state-of-the-art baselines that are designed for trip recommendation:
\textbf{TRAR} \cite{gu2020enhancing} proposes the concept of attractive routes and enhances trip recommendation with attractive routes.
\textbf{PERSTOUR} \cite{lim2015personalized} personalizes the duration time for each user based on their preferences and generates trips to maximize user preference. 
\textbf{C-ILP} \cite{jia2019joint} learns a context-aware POI embedding by integrating POI co-occurrences, user preferences and POI popularity, and transforms the problem to an integer linear programming problem.
\zhou{For C-ILP and PERSTOUR, we first generate 100 candidates by using our proposed retrieval procedure \li{because larger candidates can not be solved in a tolerable time}.  \zjb{C-ILP and PERSTOUR both utilize} \textit{lpsolve} \cite{berkelaar2004lpsolve}, a linear programming package, to generate trips among the candidates, which follows their implementation.}
% \zhoucom{1) can we give a citation to lpsolve package? 2) why TRAR does not use recall method?}

To fully validate the effectiveness of our proposed method, we introduce some baselines designed for POI recommendation.
These baselines share the same trip generation procedure: they repeatedly choose the POI with the highest score among all the unvisited POIs until the time budget exhausts.
The scores of POIs are produced by the corresponding model in SAE-NAD and GRU4Rec while the scores are popularity(visit frequency) of POIs in POP. 
\textbf{POP} is a naive method that measures the popularity of POIs by counting the visit frequency of POIs. \textbf{SAE-NAD} \cite{ma2018point} applies a self-attentive encoder for presenting POIs and a neighbor-aware decoder for exploiting the geographical influence.
\textbf{GRU4Rec} \cite{bal2016session} models the sequential information by GRU.
The implementation and hyper-parameters will be reported in the appendix.
% \begin{itemize}
%     \item \textbf{POP} is a naive method which measures popularity of POIs by counting the visit frequency of POIs.
%     \item \textbf{SAE-NAD} \cite{ma2018point} applies a self-attentive encoder for presenting POIs and a neighbor-aware decoder for exploiting the geographical influence.
%     \item \textbf{GRU4Rec} \cite{bal2016session} models the sequential information by GRU.
% \end{itemize}
\begin{table*}[tb]
    \centering
    \begin{tabular}{cccccccccccccccccc}
    \toprule
    \multirow{2}{*}{Method} & \multicolumn{2}{c}{NYC}  & \multicolumn{2}{c}{Tokyo}  & \multicolumn{2}{c}{Beijing}  & \multicolumn{2}{c}{Chengdu}  \\
    \cmidrule(r){2-3}  \cmidrule(r){4-5}  \cmidrule(r){2-3}  \cmidrule(r){6-7}  \cmidrule(r){8-9}
    & HR  & OSP & HR  & OSP & HR  & OSP & HR  & OSP \\
    \midrule
    POP     & 0.0397 & 0.0036 & 0.0482 & 0.0128 & 0.0461 & 0.0102 & 0.0483 & 0.0131 \\
    SAE-NAD & 0.0119  & 0.0001  & 0.0875 & 0.0003 & 0.1345 & 0.0005 & 0.1220 & 0.0005 \\
    GRU4Rec & 0.0181 & 0.0012 & 0.0276 & 0.0018 & 0.0307 & 0.0020 & 0.0206 & 0.0015 \\
    TRAR    & 0.0047 & 0.0020 & 0.0010 & 0.0001  & 0.0013  & 0.0001 & 0.0027 & 0.0001 \\
    PERSTOUR & 0.0075 & 0.0013 & 0.0197 & 0.0048 & 0.0134 & 0.0021 &0.0145 & 0.0024 \\
    % MARKOV & 0.1399 & 0.0532 & 0.1216 & 0.0427 & 0.1151 & 0.0189 &0.1190 & 0.0206 \\
    C-ILP   & 0.0449 & 0.0031 & 0.0241 & 0.0012 & 0.0160 & 0.0001 & 0.0175 & 0.0003\\
    ANT & \textbf{0.2103} & \textbf{0.1154} & \textbf{0.1922} & \textbf{0.1232} & \textbf{0.1610} & \textbf{0.0388} &\textbf{ 0.1348} & \textbf{0.0349}  \\
    \bottomrule
    \end{tabular}
    \caption{Comparison with baselines.}
    \label{tab-result}
    \vspace{-2mm}
\end{table*}

\subsubsection{Evaluation Metrics}
A trip is determined by the POIs that compose the trip and the order of POIs in the trip.
We evaluate these two aspects by Hit Ratio and Order-aware Precision \cite{huang2019dynamic} respectively.
\zhou{These two metrics are popularly used for trip recommendation (and planning) in previous studies.}

\textbf{Hit Ratio}. Hit Ratio (HR) is a recall-based metric, which measures how many POIs in the real trip are covered in the planned trip except the start POI:
$
    HR = \frac{ |S \cap \hat{S}| - 1 }{|\hat{S}| - 1}
$.

\textbf{Order-aware Sequence Precision} \cite{huang2019dynamic}. Order-aware Sequence Precision \zjb{(OSP)} measures the order precision of overlapped part between the real trip and the generated trip except the start POI, which is defined as:
$OSP = M / B$
where B is the number of all POI pairs in the overlapped part and M is the number of the pairs that contain the correct order. \li{We give an example in the appendix.}
% For instance, if the real trip is $l_0 \rightarrow l_1 \rightarrow l_2 \rightarrow l_3 \rightarrow l_4$ and the planned trip is $l_0 \rightarrow l_2 \rightarrow l_5 \rightarrow l_1 \rightarrow l_4$,
% it can be calculated that $HR = \frac{4-1} {5-1} = 0.75$.
% As for OSP, the overlapped part is $(l_2, l_1, l_4)$ and all the ordered POI pairs in the overlapped is $\{l_2 \Rightarrow l_1, l_2\Rightarrow l_4, l_1\Rightarrow l_4\}$, i.e., $B=3$. And $\{l_2 \Rightarrow l_1, l_2 \Rightarrow l_4 \}$ has the correct order as the real trip, so $M =2$ and $OSP = 0.67$.

\subsection{Experimental Results}
%\subsubsection{Overall Performance}
\subsubsection{Effectiveness}
Table \ref{tab-result} shows the performance \zhou{under} HR and OSP \zhou{metrics} on the four datasets with respect to different methods.
% The performance comparison of our proposed method and baselines are summarized in the Table \ref{tab-result}.
% Note that PERSTOUR \cite{lim2015personalized} and C-ILP \cite{jia2019joint} use the same \textit{lpsolve} linear programming package according to their implementation.
It can be observed that our proposed method consistently outperforms all the baselines \zhou{with a significant margin} on all the four datasets, especially on OSP, \zhou{which demonstrates that our method can recommend high-quality trips.}
PERSTOUR and CILP are both based on integer linear programming, which restricts them to respond in real time when the number of locations is large and affects their performance.
\li{TRAR is ill-behaved because modeling users and POIs only in category space are not enough to extract informative features to recommend reasonable trips.}
SAE-NAD is a strong baseline with \zhou{good} performance on HR while it performs poorly on OSP, which validates that conventional POI recommendation methods are not capable of \zhou{being extended to support trip recommendation directly.}

Due to the page limit, we omit the results on Beijing and Chengdu \zjb{if without specification} in the following analysis and the conclusions are similar on these two datasets, which \zhou{can be found in the appendix.}
% \jiang{ will be discussed in detail in the appendix.}

\subsubsection{Efficiency}
\begin{figure}[t]
    \centering
    \includegraphics[width=0.49\columnwidth]{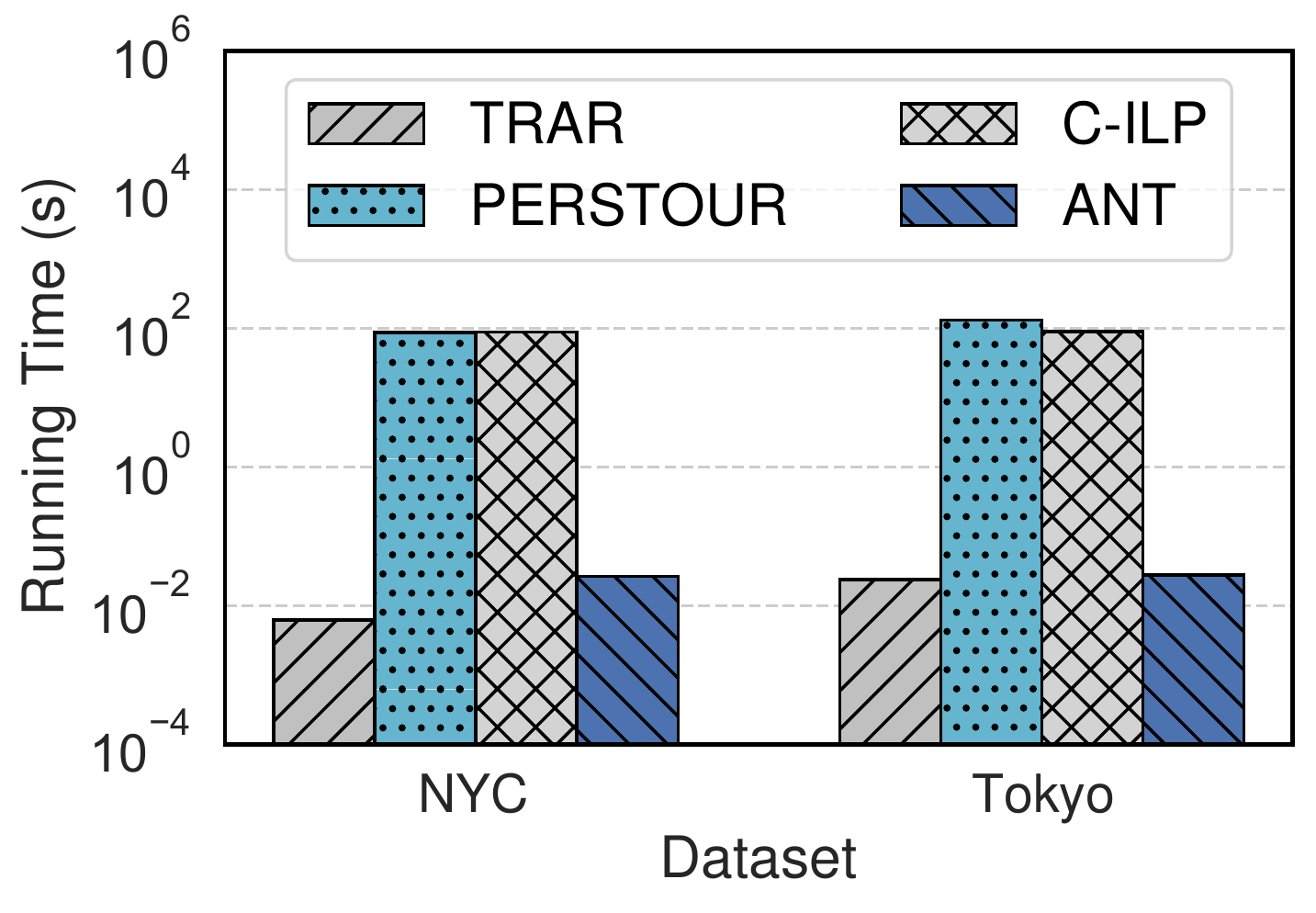}
    \includegraphics[width=0.49\columnwidth]{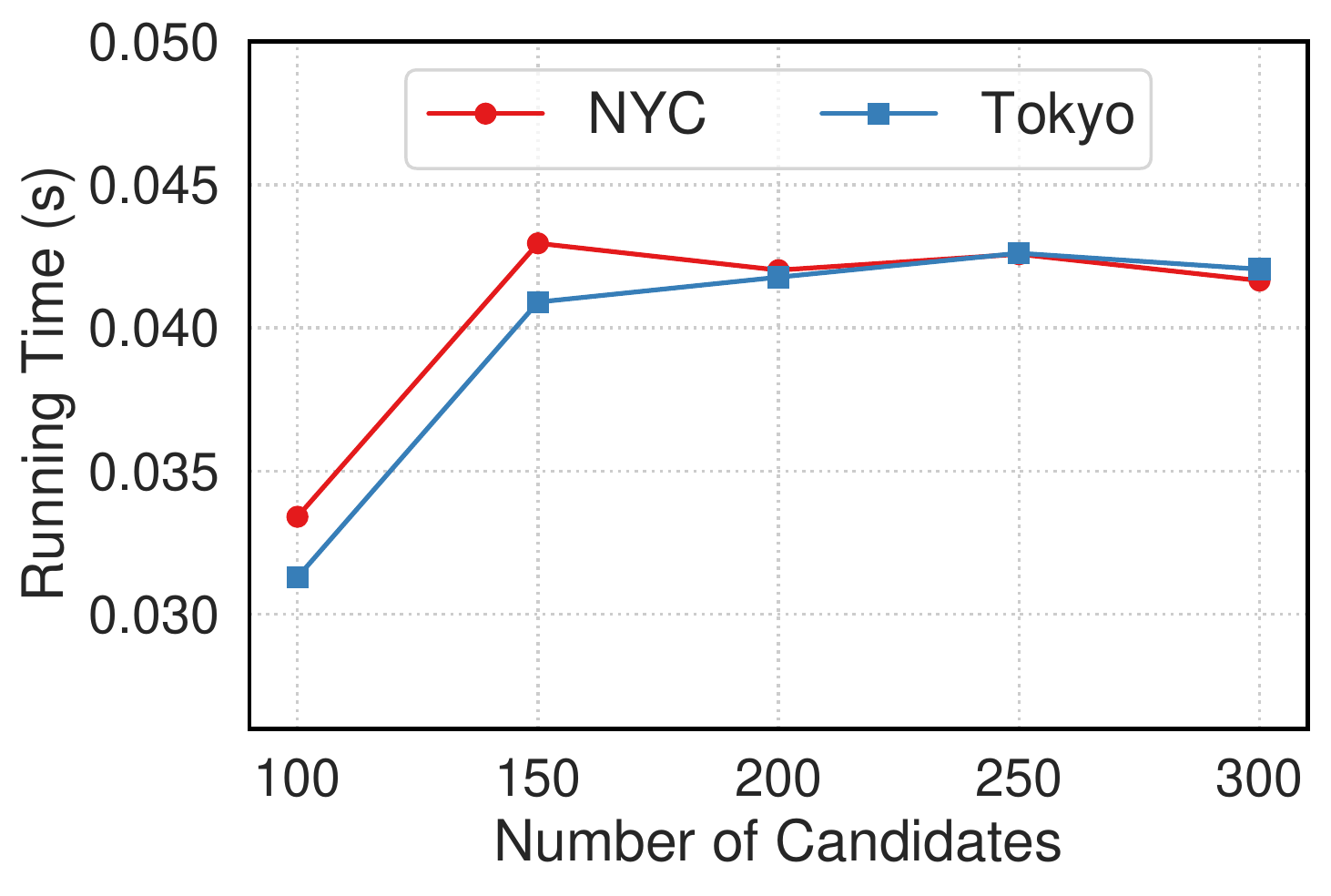}
    \vspace{-2mm}
    \caption{Running time compared with baselines and running time on different numbers of candidates.}
    \label{fig:time}
    \vspace{-2mm}
\end{figure}

\zhou{Besides the high prediction accuracy, another advantage of our framework is its good efficiency that \li{is} investigated in this section.}
We compare the running time of ANT with trip recommendation baselines, \zhou{i.e. TRAR, C-ILP and PERSTOUR}. 
% \zhoucom{whya not TRAR?}
Even though ANT can be parallelized,
\zhou{for fair comparison} we make ANT generate trips serially and we run all four methods on the same CPU device (Intel 6258R).
The average running time of C-ILP and PERSTOUR \li{on an instance }both exceed one minute, while the average running time of ANT is less than 45 ms, which
\zhou{demonstrates} the superiority of our model in efficiency compared to traditional CP-based models.
Even though TRAR is faster than ANT with the help of the greedy algorithm but TRAR's performance is much worse than ANT, even worse than PERSTOUR and C-ILP.
To further validate ANT's availability to scale up to various numbers of candidates, we run ANT conditioned on \zjb{varying} numbers of candidates and show the result in Figure \ref{fig:time}, \zhou{which shows that the inference time of ANT is relatively stable with \zjb{different} numbers of candidates.}
% \zhoucom{explain why TRAR is fast, and explain its peroformance is worse than ours, and worse than PERSTOUR and C-ILP.}

\begin{figure}[t]
    \centering
    \includegraphics[width=0.49\columnwidth]{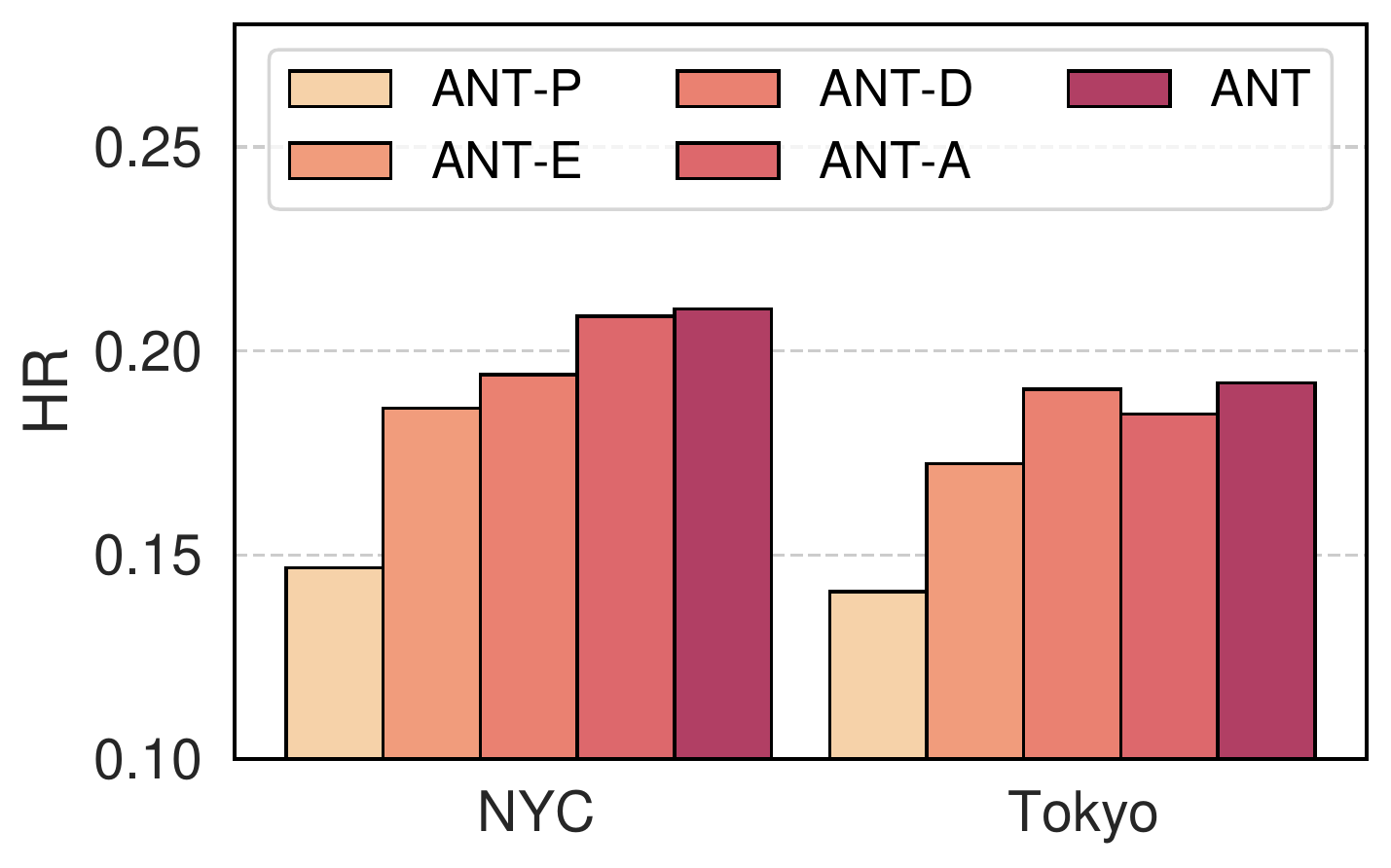}
    \includegraphics[width=0.49\columnwidth]{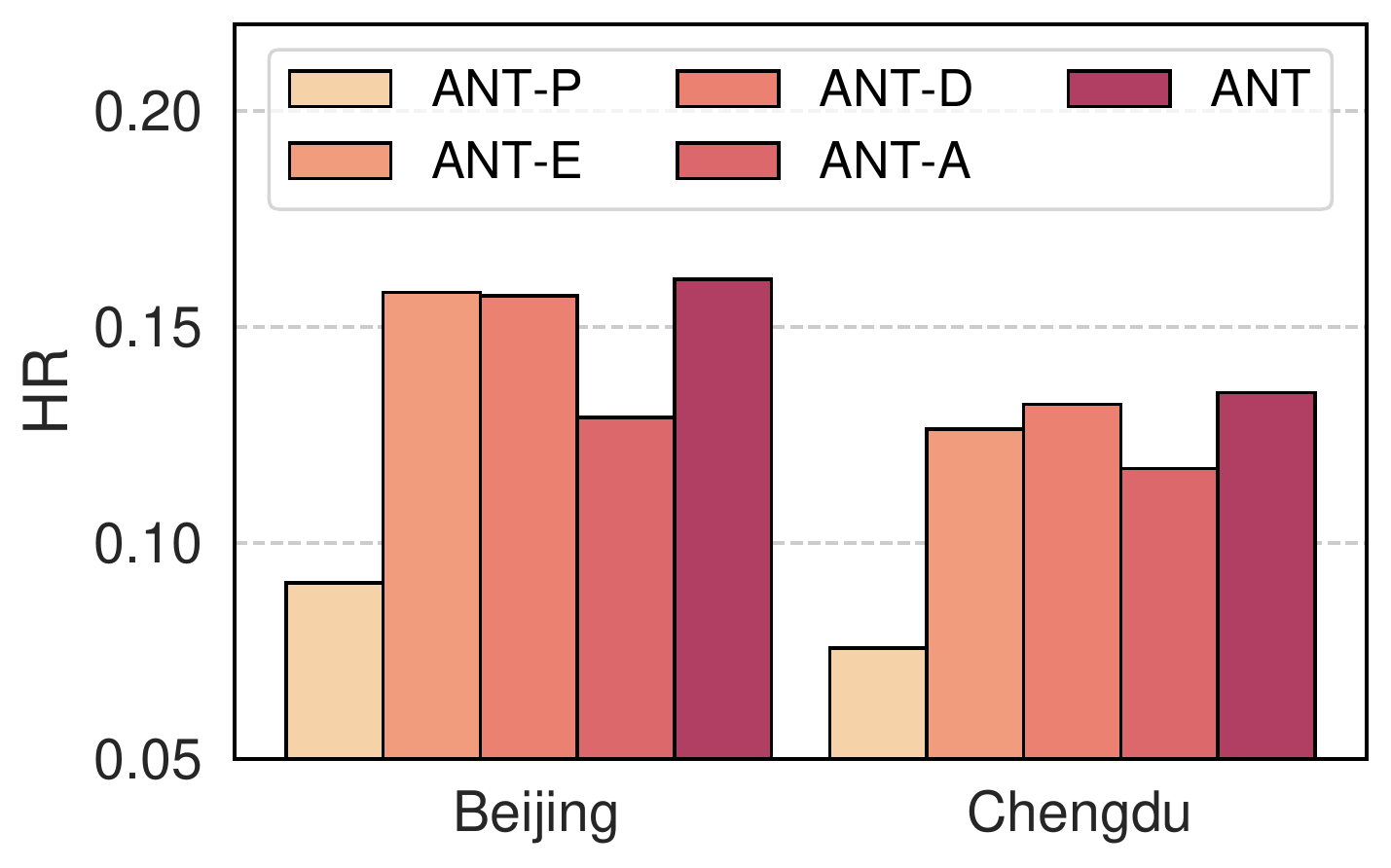}
    \vspace{-2mm}
    \caption{\zjb{Ablation study} of each component.}
    \label{fig:ablation}
    \vspace{-2mm}
\end{figure}
\subsubsection{Ablation Study}

\begin{figure}[t]
    \centering
    \includegraphics[width=0.49\columnwidth]{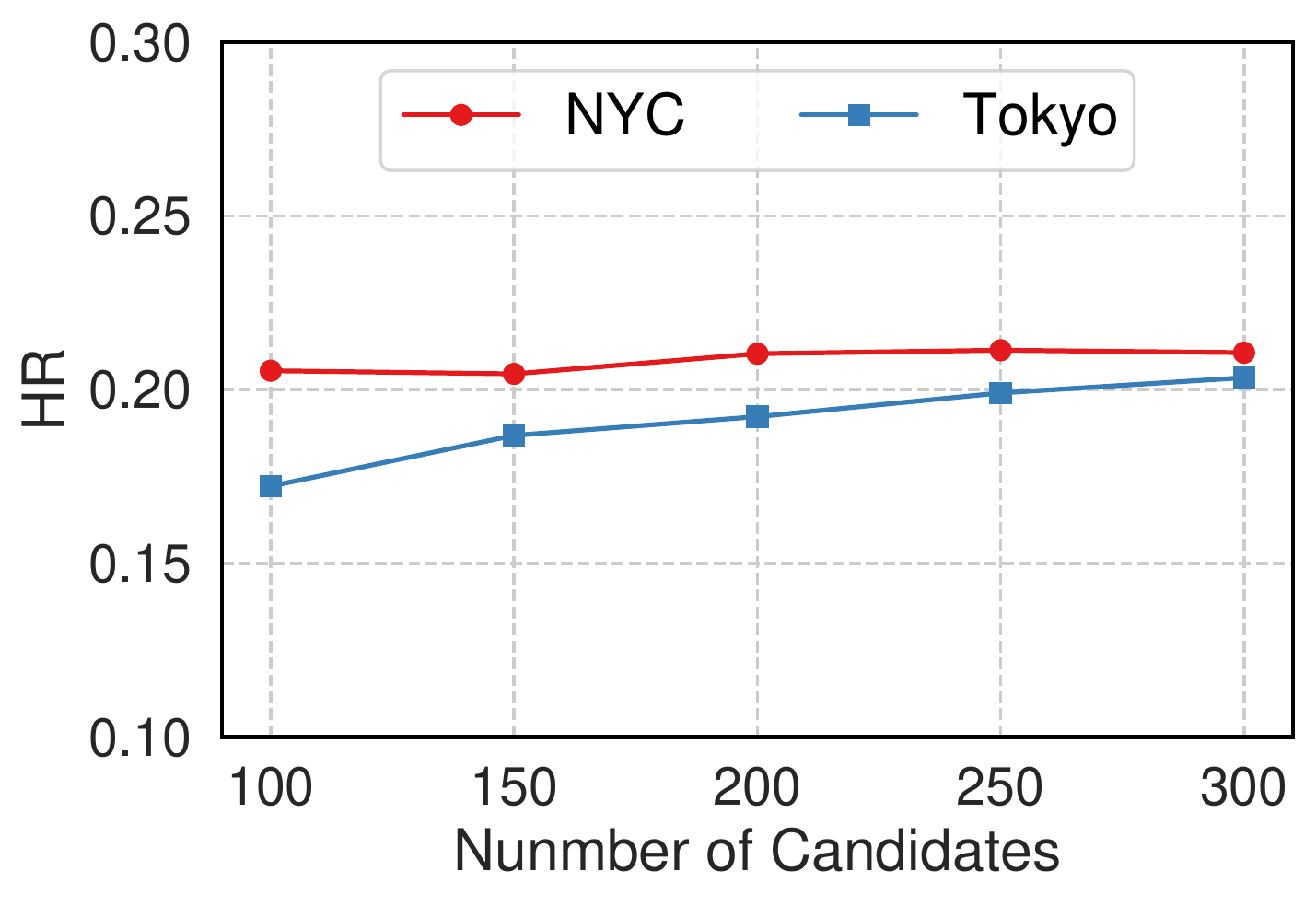}
    \includegraphics[width=0.49\columnwidth]{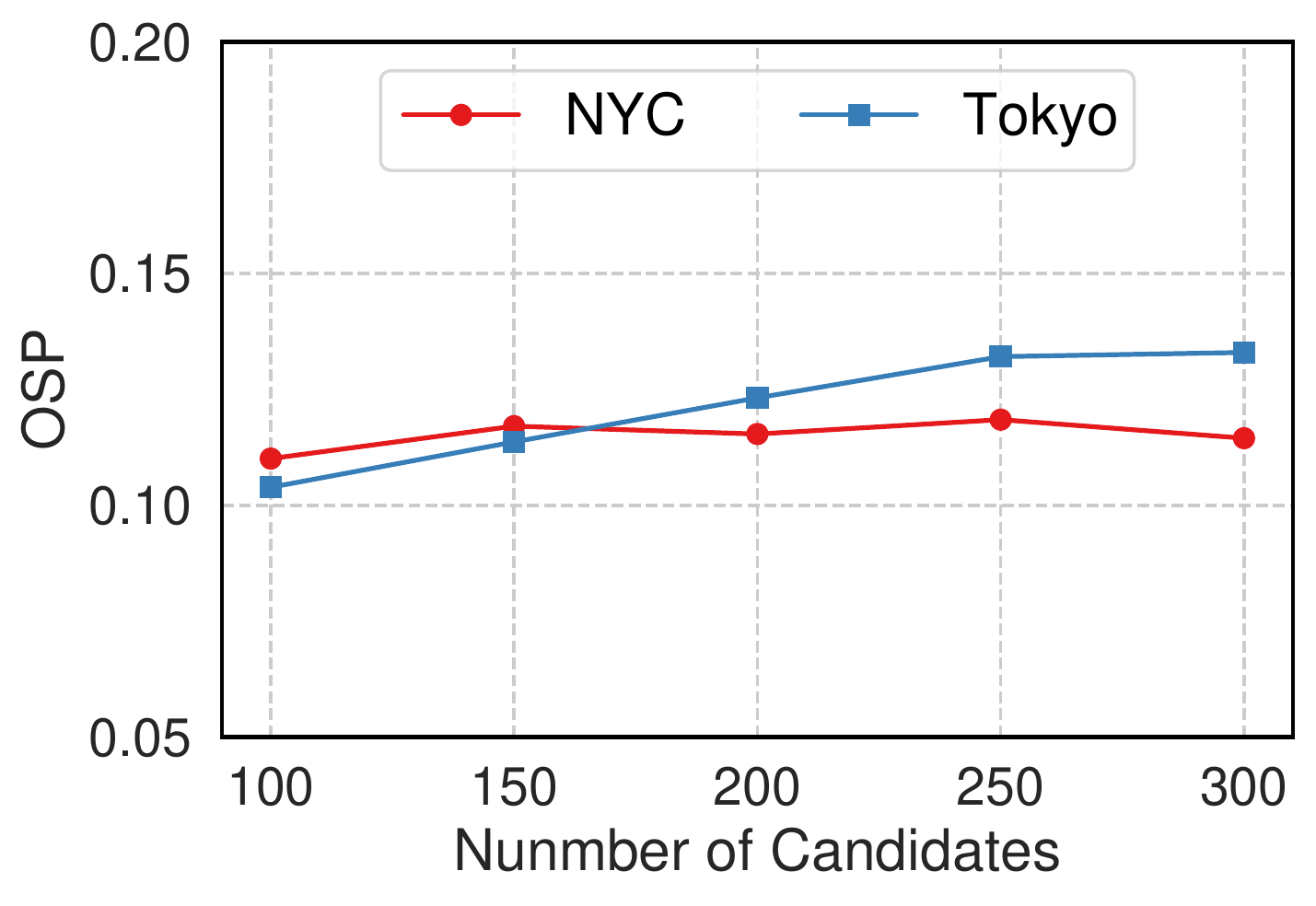}
    \caption{The impact of the number of candidates.}
    \label{fig:candidates}
    \vspace{-2mm}
\end{figure}

\zhou{To analyze the effect of each component of the ANT framework, we conduct an experimental evaluation on four variants of ANT:} ANT-E, ANT-D, ANT-A, ANT-P.
ANT-E means 
\zhou{to} remove the POI correlation module, ANT-D means the trip generation module is replaced with Pointer Networks \cite{vinyals2015pointer}, ANT-A means that we train the whole model only using learning from demonstration and ANT-P means we train the model only using the adversarial learning.
Due to the page limit, we omit the performance under OSP, which can be found in the appendix.
% \zhoucom{why not to prove the effectiveness of pre-train?}

As can be observed in Figure \ref{fig:ablation}, each component makes contributions to the final performance.
Training our model without pre-training leads to huge performance degradation, indicating the effectiveness of pre-training on stabilizing and speeding up the training process. 
Adversarial learning makes the model further improved based on pre-training.
Also, removing the POI correlation module leads to a  performance drop, indicating the necessity of \li{multi-head self-attention} to capture the POI correlation. 
% \zhoucom{by what to capture the pOI correlation?}.
And compared to Pointer Networks, the well-designed context embedding for trip recommendation also shows its superiority.
% \zhoucom{what is the effect of pre-train?}

% \zhoucom{revise the following parts later}
% \subsubsection{Parameter Sensitivity}
% We vary the number of layers in POI correlation encoding module to study the impact of the depth of the module.
% As can be observed in Figure \ref{fig:layers}, on both NYC and Tokyo, the HR is raising as the module gets deeper.
% However, the OSP in kind of sensitive to the depth on NYC.
% Thus, we set the number of layers as 6 to capture POI correlation better.

\subsubsection{Impact of Candidates}
\zhou{Here we evaluate the impact of candidates on the performance of ANT. Intuitively, when increasing the number of candidates, the target POIs have a high probability to be included in the candidate set for trip recommendation, but it also raises the difficulty to plan the correct trips. As we can see from Figure \ref{fig:candidates}, when the number of candidates is larger than 200, the prediction performance of ANT (under HR and OSP) on the NYC dataset becomes relatively stable with the number of candidates increasing. The same phenomenon can be found on the Tokyo dataset when the number of candidates is larger than 250. Therefore, the performance  is not sensitive to the number of candidates if the number is relatively large enough. In this experimental evaluation, we set the number of candidates to 200, which can be also adjusted according to different characteristics of different cities.}

% Figure \jiang{pass} shows the result and we can observe that on with the increase of layers, the performance 
% And to investigate the impact of the number of  candidates, we vary the number of candidates and report the result in Figure \jiang{pass}.
% \begin{figure}[t]
%     \centering
%     \includegraphics[width=0.495\columnwidth]{exp-fig/aaai-layers-hr.pdf}
%     \includegraphics[width=0.495\columnwidth]{exp-fig/aaai-layers-osp.pdf}
%     \caption{The impact of the number of layers.}
%     \label{fig:layers}
% \end{figure}

% To investigate our model's ability to scale up to different quantity of locations, we compare its performance and running time during inference upon different numbers of candidates.
% The result is illustrated in Figure, which shows that the running time of ANT is not sensitive to the quantity of locations.

\section{CONCLUSION}
% 将问题转换为MDP，然后用强化学习求解，为了学习好的策略，
% \jiang{In this paper, we tackle the trip recommendation problem in an end-to-end way by proposing a novel adversarial learning strategy with reinforcement learning to generate well-designed trips.
% Along this line, we devise a encoder-decoder based generator to learn a well-formed policy to select optimal POI at each time step by integrating POI correlation and contextual environment.
% The extensive results on the large-scale real-world datasets demonstrate our framework could remarkably outperform the state-of-the-art baselines both on velocity and effect.}
\zhou{In this paper, we investigated the trip recommendation problem by an end-to-end deep learning model. Along this line, we devised an encoder-decoder based trip generator to learn a well-formed policy to select the optimal POI at each time step by integrating POI correlation and contextual environment. Especially, we proposed a novel adversarial learning strategy integrating with reinforcement learning to train the trip generator. The extensive results on four large-scale real-world datasets demonstrate our framework could remarkably outperform the state-of-the-art baselines both on effectiveness and efficiency.}

\bibliography{aaai22}

\begin{thebibliography}{28}
\providecommand{\natexlab}[1]{#1}

\bibitem[{Bellman(1957)}]{bellman1957markovian}
Bellman, R. 1957.
\newblock A Markovian decision process.
\newblock \emph{Journal of mathematics and mechanics}, 6(5): 679--684.

\bibitem[{Bello et~al.(2017)Bello, Pham, Le, Norouzi, and
  Bengio}]{bello2016neural}
Bello, I.; Pham, H.; Le, Q.~V.; Norouzi, M.; and Bengio, S. 2017.
\newblock Neural Combinatorial Optimization with Reinforcement Learning.
\newblock In \emph{ICLR}.

\bibitem[{Bengio et~al.(2015)Bengio, Vinyals, Jaitly, and
  Shazeer}]{samy2015scheduled}
Bengio, S.; Vinyals, O.; Jaitly, N.; and Shazeer, N. 2015.
\newblock Scheduled Sampling for Sequence Prediction with Recurrent Neural
  Networks.
\newblock In \emph{NIPS}, 1171--1179.

\bibitem[{Berkelaar et~al.(2004)Berkelaar, Eikland, Notebaert
  et~al.}]{berkelaar2004lpsolve}
Berkelaar, M.; Eikland, K.; Notebaert, P.; et~al. 2004.
\newblock lpsolve: Open source (mixed-integer) linear programming system.
\newblock \emph{Eindhoven U. of Technology}, 63.

\bibitem[{Chen, Ong, and Xie(2016)}]{chen2016learning}
Chen, D.; Ong, C.~S.; and Xie, L. 2016.
\newblock Learning points and routes to recommend trajectories.
\newblock In \emph{CIKM}, 2227--2232.

\bibitem[{Cho et~al.(2014)Cho, Van~Merri{\"e}nboer, Gulcehre, Bahdanau,
  Bougares, Schwenk, and Bengio}]{cho2014learning}
Cho, K.; Van~Merri{\"e}nboer, B.; Gulcehre, C.; Bahdanau, D.; Bougares, F.;
  Schwenk, H.; and Bengio, Y. 2014.
\newblock Learning Phrase Representations using {RNN} Encoder-Decoder for
  Statistical Machine Translation.
\newblock In \emph{EMNLP}, 1724--1734. {ACL}.

\bibitem[{Golden, Levy, and Vohra(1987)}]{golden1987orienteering}
Golden, B.~L.; Levy, L.; and Vohra, R. 1987.
\newblock The orienteering problem.
\newblock \emph{Naval Research Logistics (NRL)}, 34(3): 307--318.

\bibitem[{Goodfellow et~al.(2014)Goodfellow, Pouget{-}Abadie, Mirza, Xu,
  Warde{-}Farley, Ozair, Courville, and Bengio}]{goodfellow2014generative}
Goodfellow, I.~J.; Pouget{-}Abadie, J.; Mirza, M.; Xu, B.; Warde{-}Farley, D.;
  Ozair, S.; Courville, A.~C.; and Bengio, Y. 2014.
\newblock Generative Adversarial Nets.
\newblock In \emph{NIPS}, 2672--2680.

\bibitem[{Gu et~al.(2020)Gu, Song, Jiang, Wang, and Liu}]{gu2020enhancing}
Gu, J.; Song, C.; Jiang, W.; Wang, X.; and Liu, M. 2020.
\newblock Enhancing Personalized Trip Recommendation with Attractive Routes.
\newblock In \emph{AAAI}, 662--669.

\bibitem[{Hao et~al.(2016)Hao, Zhou, Cheng, Huang, and Wu}]{hao2016user}
Hao, T.; Zhou, J.; Cheng, Y.; Huang, L.; and Wu, H. 2016.
\newblock User identification in cyber-physical space: a case study on mobile
  query logs and trajectories.
\newblock In \emph{SIGSPATIAL}, 1--4.

\bibitem[{Hao et~al.(2020)Hao, Zhou, Cheng, Huang, and Wu}]{hao2020unified}
Hao, T.; Zhou, J.; Cheng, Y.; Huang, L.; and Wu, H. 2020.
\newblock A unified framework for user identification across online and offline
  data.
\newblock \emph{IEEE Transactions on Knowledge and Data Engineering}.

\bibitem[{He, Qi, and Ramamohanarao(2019)}]{jia2019joint}
He, J.; Qi, J.; and Ramamohanarao, K. 2019.
\newblock A Joint Context-Aware Embedding for Trip Recommendations.
\newblock In \emph{ICDE}, 292--303. {IEEE}.

\bibitem[{Hidasi et~al.(2016)Hidasi, Karatzoglou, Baltrunas, and
  Tikk}]{bal2016session}
Hidasi, B.; Karatzoglou, A.; Baltrunas, L.; and Tikk, D. 2016.
\newblock Session-based Recommendations with Recurrent Neural Networks.
\newblock In Bengio, Y.; and LeCun, Y., eds., \emph{ICLR}.

\bibitem[{Huang et~al.(2019)Huang, Liu, Chen, and Jia}]{huang2019dynamic}
Huang, J.; Liu, Y.; Chen, Y.; and Jia, C. 2019.
\newblock Dynamic Recommendation of POI Sequence Responding to Historical
  Trajectory.
\newblock \emph{ISPRS International Journal of Geo-Information}, 8(10).

\bibitem[{Kool, van Hoof, and Welling(2019)}]{kool2018attention}
Kool, W.; van Hoof, H.; and Welling, M. 2019.
\newblock Attention, Learn to Solve Routing Problems!
\newblock In \emph{ICLR}.

\bibitem[{Li et~al.(2017)Li, Monroe, Shi, Jean, Ritter, and
  Jurafsky}]{li2017adversarial}
Li, J.; Monroe, W.; Shi, T.; Jean, S.; Ritter, A.; and Jurafsky, D. 2017.
\newblock Adversarial Learning for Neural Dialogue Generation.
\newblock In \emph{EMNLP}, 2157--2169. Association for Computational
  Linguistics.

\bibitem[{Li et~al.(2015)Li, Cong, Li, Pham, and Krishnaswamy}]{li2015rank}
Li, X.; Cong, G.; Li, X.-L.; Pham, T.-A.~N.; and Krishnaswamy, S. 2015.
\newblock Rank-geofm: A ranking based geographical factorization method for
  point of interest recommendation.
\newblock In \emph{SIGIR}, 433--442.

\bibitem[{Lian et~al.(2014)Lian, Zhao, Xie, Sun, Chen, and Rui}]{lian2014geomf}
Lian, D.; Zhao, C.; Xie, X.; Sun, G.; Chen, E.; and Rui, Y. 2014.
\newblock GeoMF: joint geographical modeling and matrix factorization for
  point-of-interest recommendation.
\newblock In \emph{SIGKDD}, 831--840.

\bibitem[{Lim et~al.(2015)Lim, Chan, Leckie, and
  Karunasekera}]{lim2015personalized}
Lim, K.~H.; Chan, J.; Leckie, C.; and Karunasekera, S. 2015.
\newblock Personalized Tour Recommendation Based on User Interests and Points
  of Interest Visit Durations.
\newblock In \emph{AAAI}, IJCAI'15, 1778–1784. AAAI Press.

\bibitem[{Luo et~al.(2020)Luo, Zhou, Bao, Li, Culpepper, Ying, Liu, and
  Xiong}]{luo2020spatial}
Luo, H.; Zhou, J.; Bao, Z.; Li, S.; Culpepper, J.~S.; Ying, H.; Liu, H.; and
  Xiong, H. 2020.
\newblock Spatial object recommendation with hints: When spatial granularity
  matters.
\newblock In \emph{SIGIR}, 781--790.

\bibitem[{Ma et~al.(2018)Ma, Zhang, Wang, and Liu}]{ma2018point}
Ma, C.; Zhang, Y.; Wang, Q.; and Liu, X. 2018.
\newblock Point-of-interest recommendation: Exploiting self-attentive
  autoencoders with neighbor-aware influence.
\newblock In \emph{CIKM}, 697--706.

\bibitem[{Silver, Bagnell, and Stentz(2010)}]{silver2010learning}
Silver, D.; Bagnell, J.~A.; and Stentz, A. 2010.
\newblock Learning from demonstration for autonomous navigation in complex
  unstructured terrain.
\newblock \emph{The International Journal of Robotics Research}, 29(12):
  1565--1592.

\bibitem[{Vaswani et~al.(2017)Vaswani, Shazeer, Parmar, Uszkoreit, Jones,
  Gomez, Kaiser, and Polosukhin}]{vaswani2017attention}
Vaswani, A.; Shazeer, N.; Parmar, N.; Uszkoreit, J.; Jones, L.; Gomez, A.~N.;
  Kaiser, L.~u.; and Polosukhin, I. 2017.
\newblock Attention is All you Need.
\newblock In \emph{NIPS}, volume~30, 5998--6008.

\bibitem[{Vinyals, Fortunato, and Jaitly(2015)}]{vinyals2015pointer}
Vinyals, O.; Fortunato, M.; and Jaitly, N. 2015.
\newblock Pointer Networks.
\newblock In \emph{NIPS}, 2692--2700.

\bibitem[{Williams(1992)}]{williams1992simple}
Williams, R.~J. 1992.
\newblock Simple statistical gradient-following algorithms for connectionist
  reinforcement learning.
\newblock \emph{Machine learning}, 8(3-4): 229--256.

\bibitem[{Yang et~al.(2017)Yang, Bai, Zhang, Yuan, and Han}]{yang2017bridging}
Yang, C.; Bai, L.; Zhang, C.; Yuan, Q.; and Han, J. 2017.
\newblock Bridging Collaborative Filtering and Semi-Supervised Learning: A
  Neural Approach for POI Recommendation.
\newblock In \emph{SIGKDD}, 1245–1254. Association for Computing Machinery.

\bibitem[{Yang et~al.(2015)Yang, Zhang, Zheng, and Yu}]{yang2014modeling}
Yang, D.; Zhang, D.; Zheng, V.~W.; and Yu, Z. 2015.
\newblock Modeling User Activity Preference by Leveraging User Spatial Temporal
  Characteristics in LBSNs.
\newblock \emph{IEEE Transactions on Systems, Man, and Cybernetics: Systems},
  45(1): 129--142.

\bibitem[{Yu et~al.(2017)Yu, Zhang, Wang, and Yu}]{yu2017seqgan}
Yu, L.; Zhang, W.; Wang, J.; and Yu, Y. 2017.
\newblock SeqGAN: Sequence Generative Adversarial Nets with Policy Gradient.
\newblock In \emph{AAAI}, 2852--2858. {AAAI} Press.

\end{thebibliography}
\clearpage
\appendix
\section{APPENDIX}
In this section, we first introduce the training procedure for ANT.
Then we give an example of evaluation metrics and introduce implementation, data pre-process and the parameter setting.
Finally, we represent the rest of experimental results, which are omitted in the experiment section.
\subsection{Algorithm}
\subsubsection{Training Algorithm for ANT}
We represent the training procedure for ANT in Algorithm \ref{alg:ANT}.

\begin{algorithm}
\caption{Training Procedure for ANT}
\label{alg:ANT}
\KwIn{Traing set $\mathcal{D}$}
\KwOut{Trained model parameters $\theta$ of the generator}
Initialize $G_\theta$, $D_\phi$ with random parameters $\theta$, $\phi$\;
Generate trips using $G_\theta$ for pre-training $D_\phi$\;
Pre-train $D_\phi$ via softmax loss function\;
Pre-train $G_\theta$ via learning from demonstration\;
\For{n\_epoches}{
\For{m\_batches}{
Generate trips by using $G_\theta$\;
Update $D_\phi$ via softmax loss function\;
Update $G_\theta$ via adversarial loss\;
Update $G_\theta$ via supervised loss function\;
\tcp{Teacher forcing}
}
}
\end{algorithm}

\subsection{Experiment Details}
\subsubsection{An Example of Evaluation Metrics}
Here we give an example about HR and OSP.
If the real trip is $l_0 \rightarrow l_1 \rightarrow l_2 \rightarrow l_3 \rightarrow l_4$ and the recommended trip is $l_0 \rightarrow l_2 \rightarrow l_5 \rightarrow l_1 \rightarrow l_4$,
it can be calculated that $HR = \frac{4-1} {5-1} = 0.75$.
As for OSP, the overlapped part is $(l_2, l_1, l_4)$ and all the ordered POI pairs in the overlapped part is $\{l_2 \Rightarrow l_1, l_2\Rightarrow l_4, l_1\Rightarrow l_4\}$, i.e., $B=3$. And $\{l_2 \Rightarrow l_1, l_2 \Rightarrow l_4 \}$ has the correct order as the real trip, so $M =2$ and $OSP = 0.67$.

\subsubsection{Data Pre-process}
The raw Foursquare dataset does not include the departure timestamp so we estimate the departure timestamp for check-ins.
As for the successive check-ins in a trip, we use the arrival timestamp on the next POI as the departure timestamp for the current POI.
Specially, for the last check-in in the trip, we set the departure timestamp as 30 minutes after the arrival timestamp.
The Map dataset already includes the full information that we need for trip recommendation so we don't pre-process it.

\subsubsection{Implementation}
We implement ANT in PyTorch and the model is trained on Tesla V100 GPU with running environment of Python 3.7, PyTorch 1.8.1 and CUDA 11.0.

\subsubsection{Parameter Setting}
We set embedding dimensions of user, POI, and category as 256, 256 and 32 respectively.
For the encoder, the dimension of multi-head self-attention is 256, the number of attention heads is 8, the inner-layer dimension of the feed-forward sublayer is 256, and we stack 6 attention layers in the encoder.
For the decoder, we set the number of attention heads as 8 and the dimension of attention is 256.
For the discriminator, we set the dimension of the hidden state of GRU as 256, the dimensions of inner layers in the feed-forward network are 32 and 2.
As for training, we set batch size as 512.
We use Adam optimizer to train our whole framework with a learning rate of 0.0001 in the pre-training stage and 0.00001 in the adversarial learning stage.

\subsection{Experimental Results}
\subsubsection{Efficiency}
\begin{figure}[t]
    \centering
    \includegraphics[width=0.49\columnwidth]{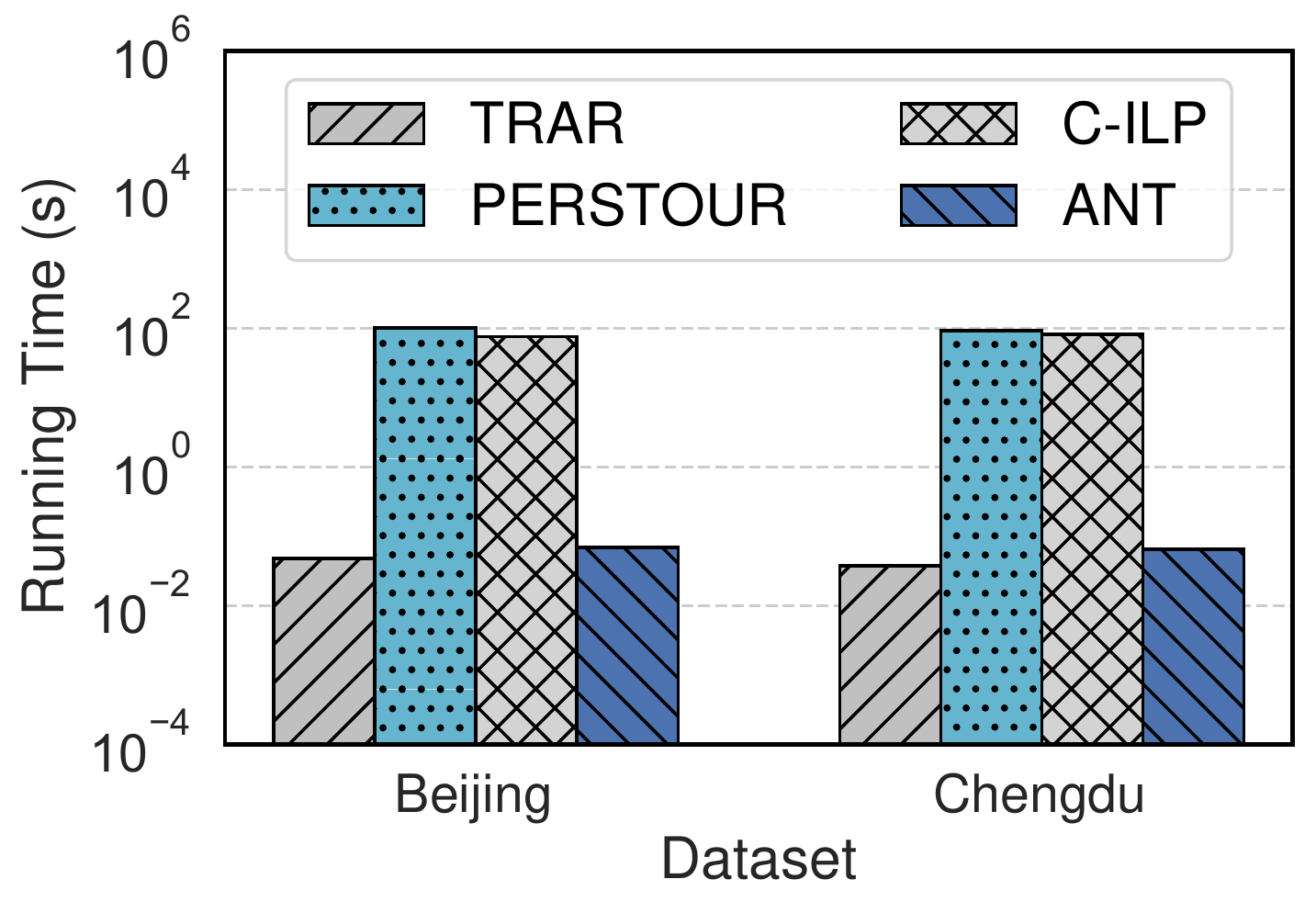}
    \includegraphics[width=0.49\columnwidth]{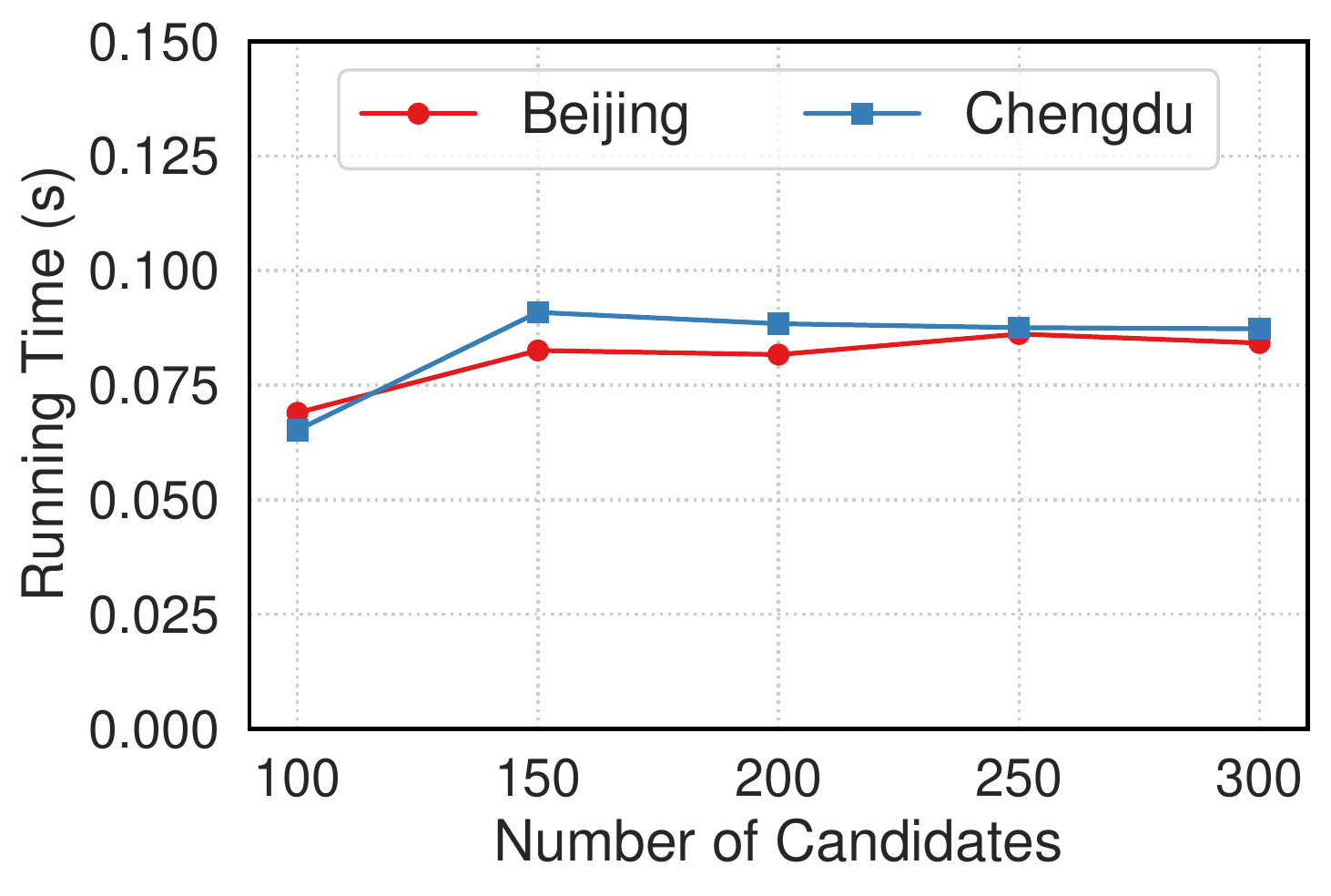}
    \caption{Running time compared with baselines and running time on different numbers of candidates.}
    \label{fig:run-time-bjcd}
\end{figure}

We compare the running time of our proposed ANT with trip recommendation baselines, i.e. TRAR, C-ILP and PERSTOUR, and the running time of ANT on different numbers of candidates.
We make ANT generate trips serially the same as described in the experiment section.
As shown in Figure \ref{fig:run-time-bjcd}, the running time of C-ILP and PERSTOUR on an instance both exceed one minute, while the running time of ANT is less than 100ms, which demonstrates the excellent efficiency of ANT.
The running time of TRAR is shorter than ANT with the help of greedy algorithm, but its performance is much worse than ANT, even worse than PERSTOUR and C-ILP.
And we also represent the running time of ANT on different numbers of candidates.
The results show that the inference time of ANT is relatively stable on different numbers of candidates.

\begin{figure}[t]
    \centering
    \includegraphics[width=0.49\columnwidth]{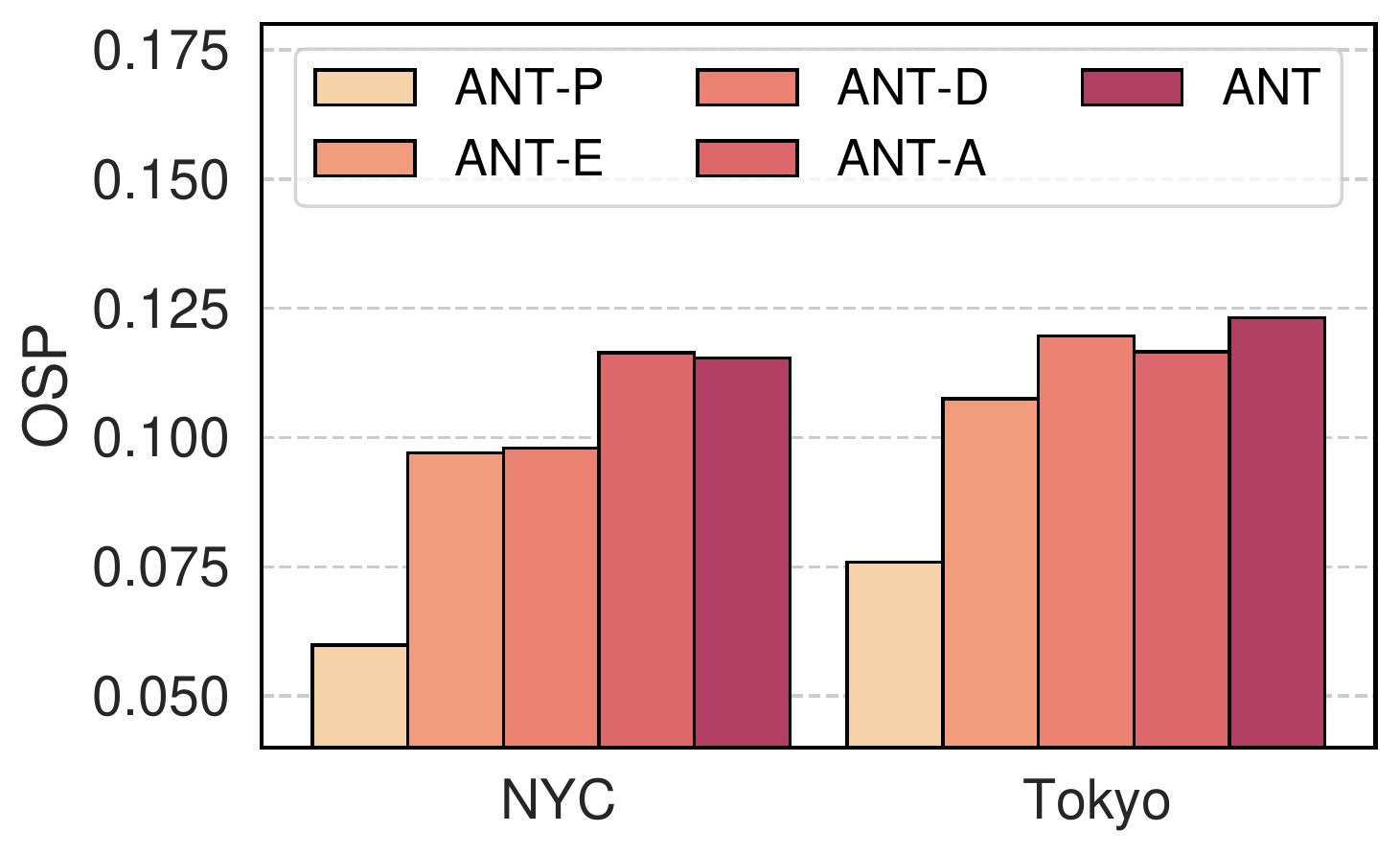}
    \includegraphics[width=0.49\columnwidth]{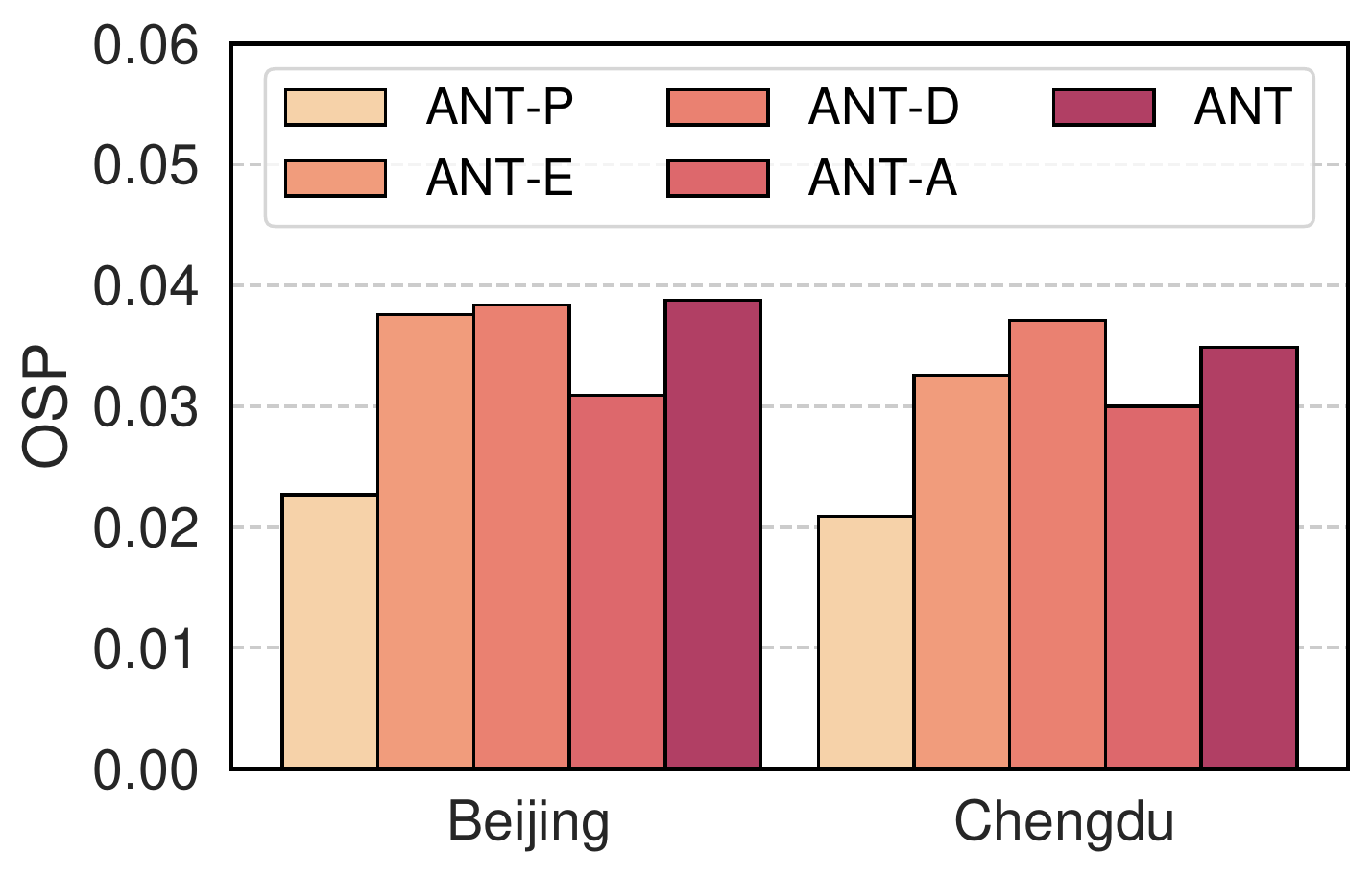}
    \caption{Ablation study of each component.}
    \label{fig:ablation-osp}
\end{figure}

\begin{figure}[t]
    \centering
    \includegraphics[width=0.49\columnwidth]{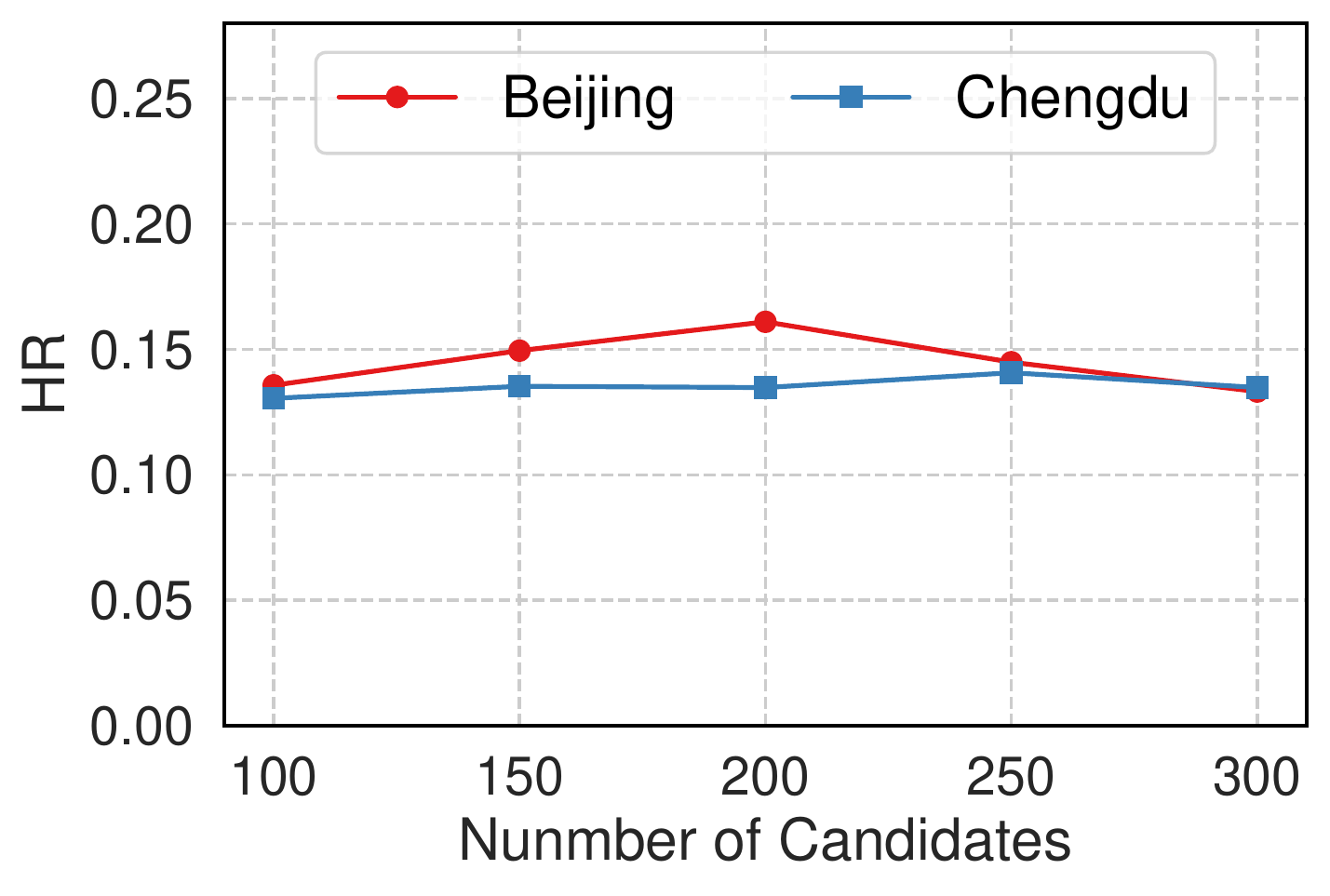}
    \includegraphics[width=0.49\columnwidth]{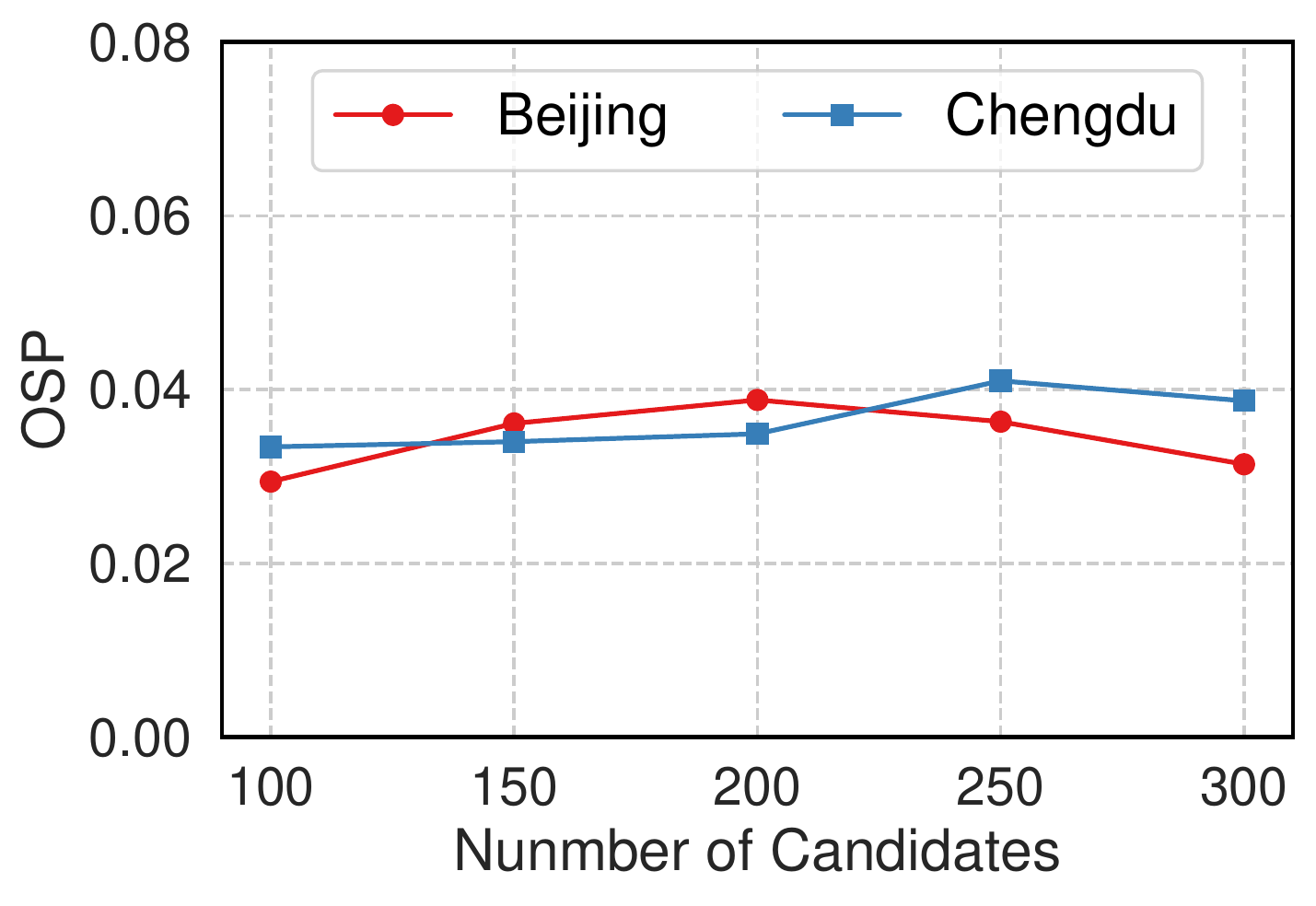}
    \caption{The impact of the number of candidates.}
    \label{fig:candidates-bjcd}
\end{figure}
\subsubsection{Ablation Study}
The performance of four variants of ANT and ANT under OSP is showed in Figure \ref{fig:ablation-osp}.
As can be observed in Figure \ref{fig:ablation-osp}, each component makes contribution to the final performance.
Training the whole framework without pre-training results in a big performance drop, which demonstrates the necessity of pre-training on stabilizing and speeding up the training process.
Based on the pre-training, adversarial learning further improves the framework.
Removing the POI correlation encoding module also makes the performance worse, indicating the effectiveness of multi-head self-attention to capture the relationships among POIs.
And our specially designed context embedding for trip recommendation also outperforms Pointer Networks.

\subsubsection{Impact of Candidates}
The performance of ANT conditioned on different numbers of candidates is illustrated in Figure \ref{fig:candidates-bjcd}.
As we can see from Figure \ref{fig:candidates-bjcd}, for Chengdu, the performance is relatively stable on different numbers of candidates under both HR and OSP.
As for Beijing, when the number of candidates is less than 200, the performance improves with increase of candidates under both HR and OSP, and when the number of candidates is more than 200, the performance deteriorates with increase of candidates.
So the proper number of candidates for Beijing is 200.
Thus, we can adjust the number of candidates according to the different characteristics of different cities.

\end{document}